\pdfoutput=1

\documentclass[11pt]{article}

\usepackage[preprint]{acl}

\usepackage{times}
\usepackage{latexsym}

\usepackage[T1]{fontenc}

\usepackage[utf8]{inputenc}

\usepackage{microtype}

\usepackage{inconsolata}

\usepackage{graphicx}
\usepackage{subcaption}

\usepackage[textsize=small]{todonotes}
\usepackage{xcolor}
\usepackage{soul}

\usepackage{xspace}
\newcommand{\proj}{\textsc{RTTC}\xspace}
\newcommand{\llamathree}{\textsc{Llama-3-8B-Inst}\xspace}
\newcommand{\llamathreeone}{\textsc{Llama-3.1-8B-Inst}\xspace}
\newcommand{\mistral}{\textsc{Mistral-7B-Inst-v0.3}\xspace}

\newcommand{\qwen}{\textsc{Qwen2.5-3B-Inst}\xspace}

\newcommand{\qwenemb}{\textsc{Qwen3-Embedding-0.6B}\xspace}
\newcommand{\skyworkeightb}{\textsc{Skywork-Reward-V2-Llama-3.1-8B}\xspace}
\newcommand{\skyworkzerosixb}{\textsc{Skywork-Reward-V2-Qwen3-0.6B}\xspace}

\newcommand{\qsc}{\textsc{QSC}\xspace}

\usepackage{amsmath} 
\usepackage{algorithmic}
\usepackage{algorithm}

\usepackage{pifont}
\usepackage{makecell}

\newcommand{\argminF}{\mathop{\mathrm{argmin}}\limits}   

\usepackage{booktabs}
\usepackage{multirow}
\usepackage{arydshln}
\usepackage{soul}

\usepackage{amsmath} 
\usepackage{pifont}
\usepackage{xcolor}
\newcommand{\distbar}[3]{%
  \raisebox{0pt}[8pt][0pt]{%
    \hspace{1pt}%
    \rule{2pt}{#1pt}%
    \hspace{0.5pt}%
    \rule{2pt}{#2pt}%
    \hspace{0.5pt}%
    \rule{2pt}{#3pt}%
  }%
}

\title{
\proj: Reward-Guided Collaborative Test-Time Compute
}

\author{J. Pablo Muñoz\thanks{Equal contribution.}\\
  Intel Labs \\
  Santa Clara, CA, USA \\
  \texttt{pablo.munoz@intel.com} \\
  \texttt{jpablomch@gmail.com}
  \And
  Jinjie Yuan$^{*}$ \\
  Intel Corporation \\
  Beijing, China \\
\texttt{jinjie.yuan@intel.com}}

\begin{document}
\maketitle

\begin{abstract}

Test-Time Compute (TTC) has emerged as a powerful paradigm for enhancing the performance of Large Language Models (LLMs) at inference, leveraging strategies such as Test-Time Training (TTT) and Retrieval-Augmented Generation (RAG).
However, the optimal adaptation strategy varies across queries, and indiscriminate application of TTC strategy incurs substantial computational overhead.
In this work, we introduce \textbf{R}eward-Guided \textbf{T}est-\textbf{T}ime \textbf{C}ompute (\textbf{\proj}), a novel framework that adaptively selects the most effective TTC strategy for each query via a pretrained reward model, maximizing downstream accuracy across diverse domains and tasks.
\proj operates in a distributed server-client architecture, retrieving relevant samples from a remote knowledge base and applying RAG or lightweight fine-tuning on client devices only when necessary.
To further mitigate redundant computation, we propose Query-State Caching, which enables the efficient reuse of historical query states at both retrieval and adaptation levels.
Extensive experiments across multiple LLMs and benchmarks demonstrate that \proj consistently achieves superior accuracy compared to vanilla RAG or TTT, validating the necessity of adaptive, reward-guided TTC selection and the potential of \proj for scalable, high-performance language model adaptation.

\end{abstract}

\section{Introduction}

\begin{figure}
  \centering
  \includegraphics[width=.45\textwidth]{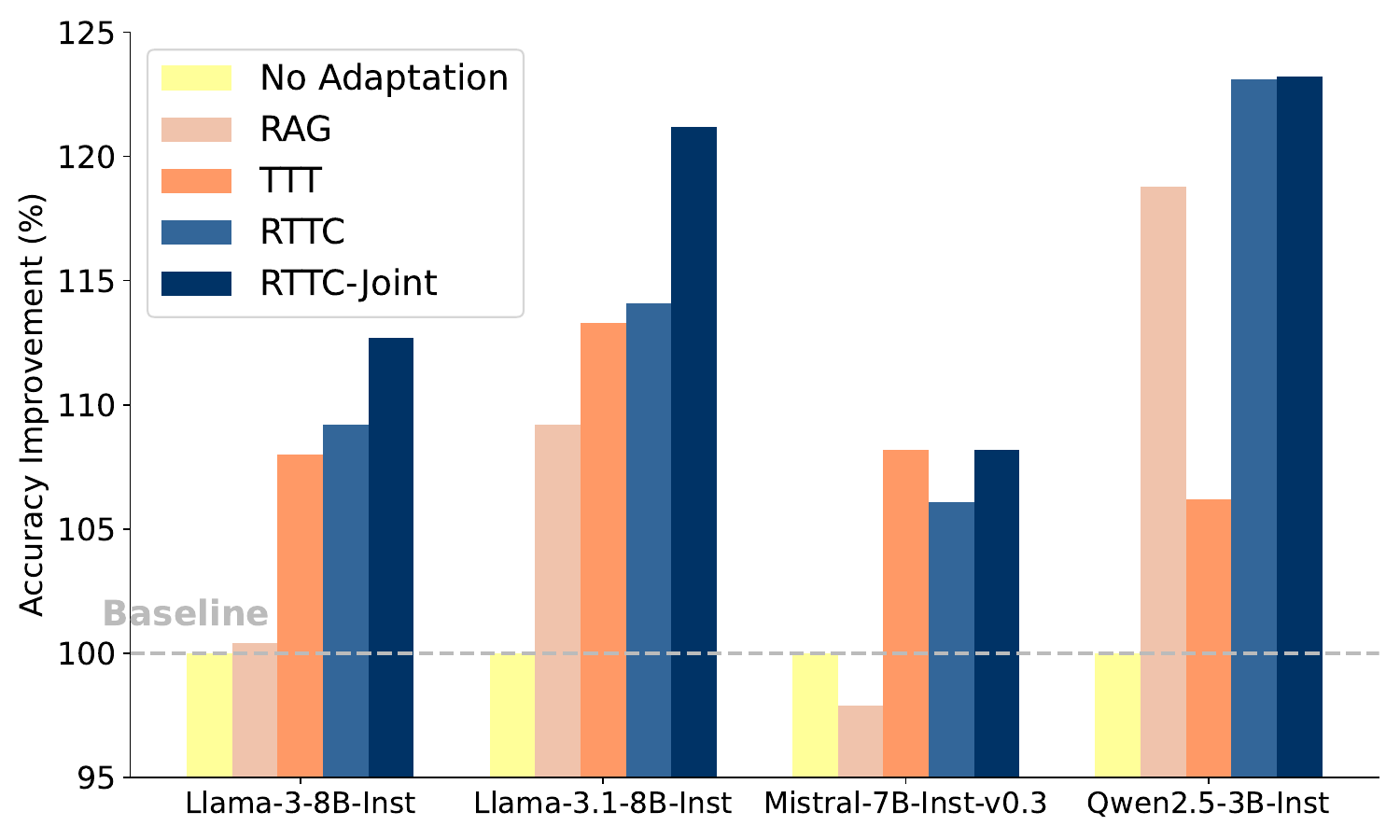}
  \caption{
  Performance overview of \proj on downstream tasks across various LLMs. “Accuracy Improvement” indicates the relative average score on five downstream tasks evaluated in \S \ref{sec:experiments}.
  }
\label{fig:results_overview}
\end{figure}

\begin{table}[ht]
\small
\setlength{\tabcolsep}{3.5pt}
\centering
\scriptsize
\begin{tabular}{llll}
\toprule
\textbf{TTC Strategy} & \textbf{Latency} & \textbf{Memory}  & \textbf{Accuracy} \\
\midrule
No Adaptation (Direct Inference) & + & + & + \\
Retrieval-Augmented Generation (RAG) & ++ & ++ & +++\\
Test-Time Training (TTT) & +++ & +++ & +++\\
\bottomrule
\end{tabular}
\caption{
Comparison of computational cost and accuracy boosting benefits for different Test-Time Compute (TTC) strategies. More plus signs indicate higher cost or increased accuracy. While RAG and TTT both provide significant accuracy improvements, each has its strengths, depending on the query and scenario.
}
\label{tab:ttc_cost}
\end{table}

Large language models have achieved remarkable performance across a wide range of tasks. However, their robustness and adaptability to new domains or distribution shifts at inference time remain open challenges. Traditionally, two major paradigms have been explored to enhance LLM performance at test time: Retrieval-Augmented Generation (RAG) \cite{lewis2020retrieval, gao2023retrieval}, which augments model input with retrieved knowledge, and Test-Time Training (TTT) \cite{sun2020test, hardt2024test, hubotter2024efficiently, snell2025scaling}, which adapts model parameters using relevant samples. Both approaches have demonstrated significant effectiveness.

\begin{figure*}
  \centering
  \includegraphics[width=\textwidth]{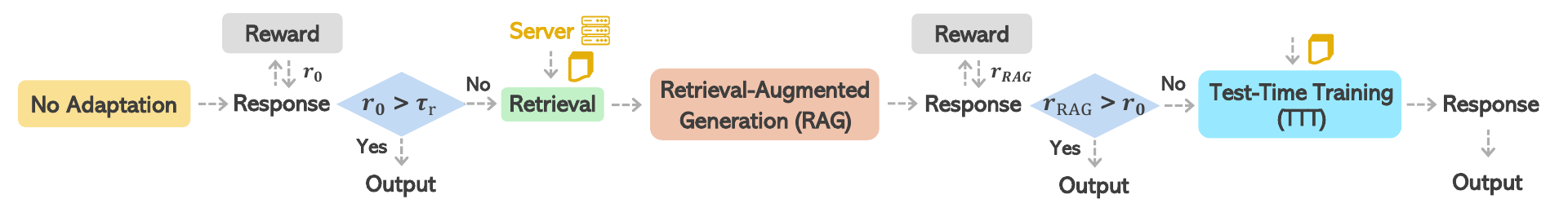}
  \caption{
  Workflow of \textbf{R}eward-guided \textbf{T}est-\textbf{T}ime \textbf{C}ompute (\textbf{\proj}). For each query, a pretrained reward model evaluates candidate responses and selects the optimal adaptation strategy (No Adaptation, RAG, or TTT). $\tau_r$ is a predefined threshold.
  }
\label{fig:rttc}
\end{figure*}

However, the practical deployment of RAG and TTT raises two concerns: accuracy and efficiency.
\textbf{Accuracy:} The effectiveness of RAG or TTT also varies across queries. Sometimes RAG outperforms TTT, and vice versa. For some inputs, the model's direct response is already sufficiently accurate, while for others, the inference stage may benefit from retrieval augmentation or adaptive fine-tuning.
\textbf{Efficiency:} Both RAG and TTT introduce significant computational overheads—RAG increases inference latency and memory usage by expanding the input context, while TTT requires additional fine-tuning steps and memory for model updates.  
Moreover, naively applying RAG or TTT to every query can lead to unnecessary computation and inefficient resource utilization. A summary of the computational cost and accuracy associated with each strategy is provided in Table~\ref{tab:ttc_cost}.
Therefore, \textbf{an adaptive approach that can dynamically select the optimal strategy at test-time for each query is crucial for maximizing performance while minimizing overhead}. 

Aiming to tackle the above challenges, we propose \textbf{R}eward-guided \textbf{T}est-\textbf{T}ime \textbf{C}omputing (\textbf{\proj}). This framework dynamically selects among three strategies for each query: No Adaptation (i.e., returning the model's response without adaptation), Retrieval-Augmented Generation (RAG), and Test-Time Training (TTT). At the core of \proj is a pretrained reward model that evaluates candidate responses and guides the system to choose the most effective adaptation strategy in a query-adaptive manner (see \autoref{fig:rttc}).
This reward-guided collaboration enables \proj to adaptively exploit the most suitable strategy for each query, achieving robust downstream performance improvements across diverse domains and tasks. Unlike prior work that statically applies either RAG or TTT, our approach introduces a principled decision-making mechanism that maximizes performance. 
Additionally, \proj also introduces the Query-State Caching (\qsc) mechanism to further optimize the test-time efficiency. \qsc leverages historical query embeddings and their associated retrieved samples or fine-tuned model state (e.g., LoRA \cite{hu2022lora}) to potentially bypass the need for repeated retrieval and fine-tuning, thus reducing computational overhead and latency. 
Overall, our main contributions are:
\begin{enumerate}
    \item We introduce \proj, a reward-guided collaborative test-time compute framework. We design an effective decision process that leverages a pre-trained reward model to adaptively choose the optimal inference strategy, enabling robust LLM adaptation.
    \item We further propose a Query-State Caching (QSC) mechanism that reuses historical query information, reducing redundant computation and latency during inference.
    \item Extensive experiments demonstrate that \proj consistently outperforms baselines, achieving higher accuracy across multiple LLMs and downstream tasks.
\end{enumerate}

\begin{figure*}
  \centering
  \includegraphics[width=.9\textwidth]{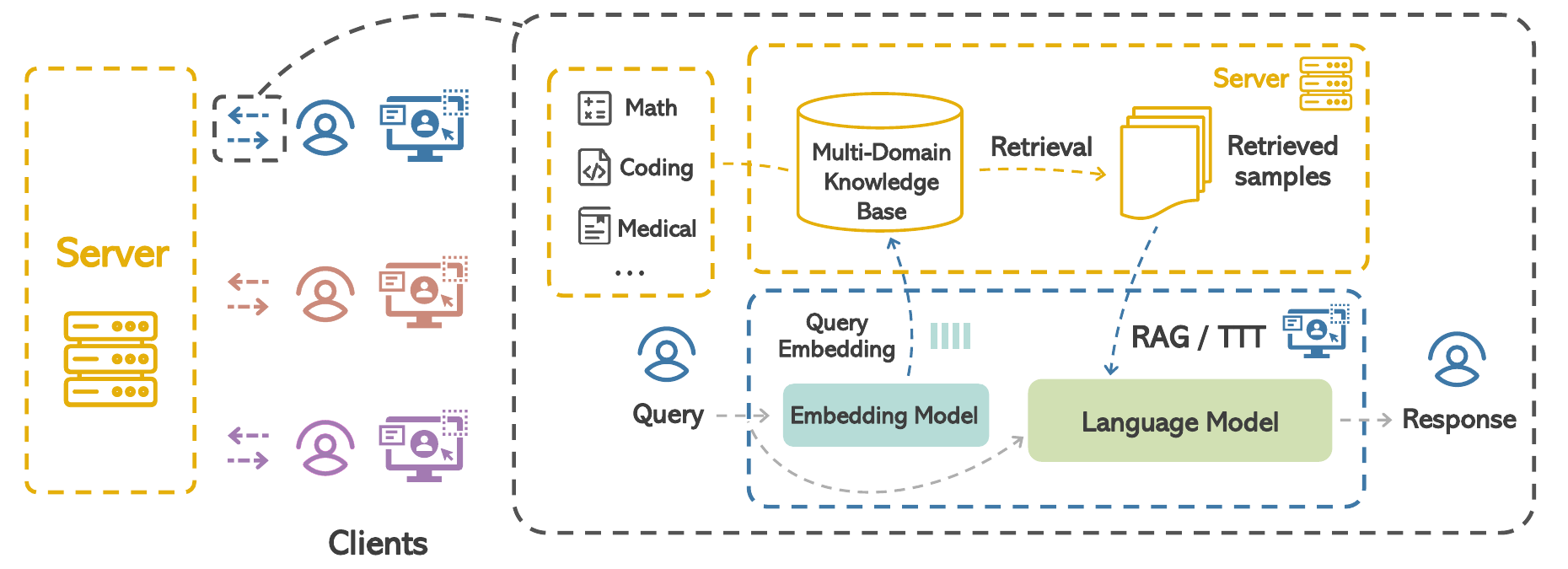}\caption{Overview of the retrieval and test-time compute stages. When the reward model determines that the LLM response without adaptation does not meet expectations, relevant samples are retrieved from a remote multi-domain knowledge base. By leveraging advanced test-time compute strategies (RAG or TTT), \proj improves model performance on client devices.}
\label{fig:retrieval}
\end{figure*}

In summary, our work is the first to unify direct inference, RAG, and TTT within a reward-guided, query-adaptive framework. The remainder of the paper is organized as follows. We discuss related work in \S \ref{sec:related_w}. Then, \S \ref{sec:system} describes the \proj system, while \S \ref{sec:experiments} presents results on various downstream tasks. Our final thoughts are in \S \ref{sec:conclusion}.

\section{Related Work}
\label{sec:related_w}

\textbf{Test-Time Compute.}
Test-time compute techniques have been proposed as an alternative to scaling model parameters for improving model performance \cite{snell2025scaling}. 
Notable strategies include \emph{Chain-of-Thought (CoT) prompting} \cite{wei2022chain} and few-shot learning \cite{NEURIPS2020_fewshot}. CoT prompting guides the model through intermediate steps to break down complex tasks, enhancing its ability to handle intricate queries. Few-shot learning helps the model adapt to new tasks with just a few examples.
Other test-time compute methods include verifying the model's results, for example, through code execution \cite{brown2025large}. 

\textbf{Retrieval-Augmented Generation (RAG).}
RAG has emerged as a prominent paradigm for enhancing model performance by incorporating external knowledge at inference time \cite{lewis2020retrieval, gao2023retrieval}. By retrieving relevant documents and augmenting the model's input, RAG enables LLMs to access up-to-date or domain-specific information beyond their pretraining corpus. This approach has demonstrated strong results across open-domain question answering, fact verification, and knowledge-intensive tasks. RAG introduces additional computational overhead due to retrieval and longer input sequences, and its effectiveness can vary depending on the quality of retrieved content and the task.

\textbf{Test-Time Training (TTT).}
TTT \cite{hardt2024test, hubotter2024efficiently, akyurek2024surprising} has proven effective for adapting models to distribution shifts by fine-tuning on retrieved samples during inference. Recent work, such as SIFT \cite{hubotter2024efficiently}, has improved retrieval strategies for TTT, while Omni-ARC \cite{ironbarArc24} leveraged TTT to achieve state-of-the-art results in the ARC-AGI challenge \cite{arcprizePrize, chollet_measure}. Despite these advances, TTT can incur significant computational and memory costs.

\textbf{System Considerations.}
In distributed settings, prior work has explored scalable retrieval and adaptation using distributed indexes and multi-server architectures to accelerate query processing over large datasets \cite{hardt2024test, douze2024faiss}. While such approaches focus on efficient data access and retrieval, our work emphasizes adaptive test-time compute and downstream task performance. \proj can flexibly incorporate these distributed retrieval techniques to further optimize efficiency when needed.

Next, we explore \proj, a system that enhances model performance via reward-guided collaborative test-time compute.

\begin{figure*}
  \centering
  \includegraphics[width=\textwidth]{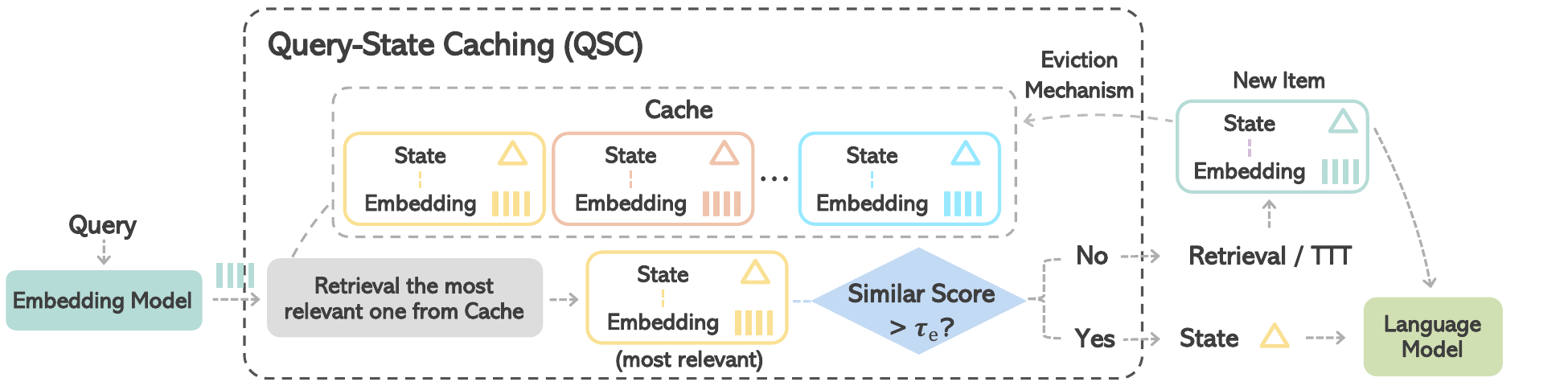}
  \caption{Overview of the Query-State Caching (\qsc) strategy for multi-turn caching and efficient test-time compute. \qsc is compatible with both RAG and TTT: for RAG, the cached state consists of retrieved samples; for TTT, the cached state stores the fine-tuned model state (e.g., LoRA adapters \cite{hu2022lora}). By managing historical query embeddings and their associated states, \qsc accelerates the test-time compute cost (retrieval and fine-tuning).}
\label{fig:lsm}
\end{figure*}

\section{\proj System}
\label{sec:system}

This section outlines the architecture and workflow of Reward-Guided Collaborative Test-Time Compute (\proj), a system designed to optimize the performance of large language models (LLMs). \proj unifies direct inference, retrieval-augmented generation (RAG), and test-time training (TTT) within a reward-driven, query-adaptive framework, leveraging a remote multi-domain knowledge base for robust adaptation. The overall workflow is shown in Figure~\ref{fig:rttc} and formally summarized in Algorithm~\ref{alg:reward_guided_pipeline}.

\subsection{Reward-Guided Test-Time Compute Pipeline}

Given an input query $x \in \mathcal{X}$, \proj orchestrates a multi-stage adaptive inference process, guided by a pretrained reward model $R$ that is utilized to dynamically select the most effective computation strategy for each query $x$. The pipeline proceeds as follows:

\paragraph{Step 1. Initial Inference and Reward Evaluation}
Upon receiving a query $x$, the LLM $M_0$ generates an initial response $\hat{y}_0 = M_0(x)$. This response is assessed by the pretrained reward model $R$, which estimates its quality $r_0 = R(x, \hat{y}_0)$. If $r_0$ surpasses a predefined threshold $\tau_r$, the system returns $\hat{y}_0$, minimizing latency and computational overhead.

\paragraph{Step 2. Retrieval of Relevant Knowledge}
When the initial response does not meet the requirements we set (i.e., $r_0 < \tau_r$), the system transitions to a retrieval phase. The query $x$ is encoded into a dense embedding $\mathbf{e}_x = E(x)$ using a shared embedding model $E$. This embedding is transmitted to a remote server hosting a multi-domain knowledge base $\mathcal{D}$, which returns a set of relevant samples $\mathcal{S}_k = \{(x_i, y_i)\}_{i=1}^k$ identified via similarity search (e.g., using FAISS \cite{douze2024faiss} and SIFT\cite{hubotter2024efficiently} algorithms). The details of distributed retrieval workflow are illustrated in Figure~\ref{fig:retrieval}.

\paragraph{Step 3. Retrieval-Augmented Generation (RAG)}
The retrieved samples $\mathcal{S}_k$ are prepended to the original query, forming an augmented input $x' = [\mathcal{S}_k; x]$. The LLM then generates a new response $\hat{y}_\text{RAG} = M_0(x')$ conditioned on this expanded context. The reward model re-evaluates the new response, yielding $r_\text{RAG} = R(x', \hat{y}_\text{RAG})$. If $r_\text{RAG} > r_0$, $\hat{y}_\text{RAG}$ is returned as the final output.

\begin{algorithm}[H]
  \caption{Reward-Guided Test-Time Compute Pipeline}
  \label{alg:reward_guided_pipeline}
  \footnotesize
  \textbf{Input:} Query $x$; LLM $M_0$; reward model $R$; embedding model $E$; knowledge base $\mathcal{D}$; threshold $\tau_r$.

  \begin{algorithmic}[1]
    \STATE \textbf{Initial Inference:} Generate $\hat{y}_0 = M_0(x)$.
    \STATE \textbf{Reward Evaluation:} Compute $r_0 = R(x, \hat{y}_0)$.
    \IF{$r_0 \geq \tau_r$}
        \STATE \textbf{Return} $\hat{y}_0$
    \ELSE
        \STATE \textbf{Retrieval:} Encode $x$ as $\mathbf{e}_x = E(x)$.
        \STATE Retrieve relevant samples $\mathcal{S}_k = \{(x_i, y_i)\}_{i=1}^k$ from $\mathcal{D}$ using similarity search.
        \STATE \textbf{RAG Inference:} Form $x' = [\mathcal{S}_k; x]$ and generate $\hat{y}_\text{RAG} = M_0(x')$.
        \STATE Compute $r_\text{RAG} = R(x, \hat{y}_\text{RAG})$.
        \IF{$r_\text{RAG} > r_0$}
            \STATE \textbf{Return} $\hat{y}_\text{RAG}$
        \ELSE
            \STATE \textbf{Test-Time Training:} Adapt $M_0$ on $\mathcal{S}_k$ to obtain $M_{\text{TTT}} = \textsc{Train}(M_0, \mathcal{S}_k)$.
            \STATE Generate final response $\hat{y}_\text{TTT} = M_{\text{TTT}}(x)$.
            \STATE \textbf{Return} $\hat{y}_\text{TTT}$
        \ENDIF
    \ENDIF
  \end{algorithmic}
\end{algorithm}

\paragraph{Step 4. Test-Time Training (TTT)}
If neither direct inference nor RAG yields a satisfactory response, \proj invokes test-time training. The same retrieved samples $\mathcal{S}_k$ are used to perform lightweight, query-specific fine-tuning of the LLM, resulting in an adapted model $\mathcal{M}_{TTT} = \textsc{Train}(\mathcal{M}_0, \mathcal{S}_k)$ via LoRA \cite{hu2022lora}. The adapted model generates the final response $\hat{y}_\text{TTT} = M_{TTT}(x)$.

\paragraph{Alternative: Joint RAG and TTT Decision}
In addition to the sequential decision process described above, \proj also supports a joint strategy wherein both RAG and TTT are executed in parallel for queries where the initial response is insufficient. The system then returns the response (either $\hat{y}_\text{RAG}$ or $\hat{y}_\text{TTT}$) with the higher reward score as determined by the reward model. While this approach can further enhance robustness by consistently selecting the best available response, it incurs additional computational and latency overhead due to the need to perform both TTT and reward evaluation for some queries. In practice, this joint strategy is optional and can be selectively enabled for scenarios where maximizing response quality is prioritized over efficiency.

\subsection{Distributed Architecture}
\proj is implemented in a server-client paradigm, where the remote server maintains the knowledge base $\mathcal{D}$ and handles retrieval, while all inference, reward evaluation, and adaptation steps are performed locally on the client device. This design can help mitigate privacy risks by keeping sensitive inference and eliminates the need to store the knowledge base locally, thereby significantly reducing memory overhead on client devices. An overview of the distributed retrieval workflow is illustrated in Figure~\ref{fig:retrieval}, which highlights how relevant samples are retrieved from a remote multi-domain knowledge base to support advanced test-time compute strategies on client devices.

\begin{algorithm}[H]
  \caption{Query-State Caching (\qsc)}
  \footnotesize
  \label{alg:qsc_unified_branches_simple}
  \textbf{Input:} \\
  \hspace{1.5em} Current query embedding $e_{x^{t}}$; set of historical query embeddings $\mathcal{Q}$;\\
  \hspace{1.5em} Reuse threshold $\tau_e$; budget $b$; similarity metric $\gamma$; eviction mechanism $\kappa$.

  \vspace{0.5em}
  \textbf{[RAG] Input:} RAG cache $\mathcal{C}_{\text{RAG}}: e_{x} \rightarrow$ retrieved samples.\\
  \textbf{[RAG] Output:} $S^{\text{RAG}}_t$ (retrieved samples).

  \begin{algorithmic}[1]
    \STATE $e_{x^{*}} = \argminF_{e_{x^{i}} \in \mathcal{Q}} \gamma(e_{x^{i}}, e_{x^{t}})$
    \IF{$\gamma(e_{x^{*}}, e_{x^{t}}) > \tau_e$}
        \STATE $S^{\text{RAG}}_t \gets \mathcal{C}_{\text{RAG}}[e_{x^{*}}]$ \COMMENT{RAG cache hit}
    \ELSE
        \STATE $S^{\text{RAG}}_t \gets \textsc{RetrieveSamples}(e_{x^{t}})$
        \IF{$|\mathcal{C}_{\text{RAG}}| \geq b$}
            \STATE $\mathcal{U} = \{e^{r}_{x^{1}} ... e^{r}_{x^{m}}\} = \kappa(\mathcal{C}_{\text{RAG}})$
            \STATE $\mathcal{C_{\text{RAG}}} \leftarrow \mathcal{C_{\text{RAG}}} \setminus \mathcal{U}$
        \ENDIF
        \STATE $\mathcal{C}_{\text{RAG}}[e_{x^{t}}] \gets S^{\text{RAG}}_t$
        \STATE $\mathcal{Q} \gets \mathcal{Q} \cup \{e_{x^{t}}\}$
    \ENDIF
  \end{algorithmic}

  \vspace{0.5em}
  \textbf{[TTT] Input:} TTT cache $\mathcal{C}_{\text{TTT}}: e_{x} \rightarrow$ trained adapters; retrieved samples $S^{\text{RAG}}_t$; initial adapter $S_0$.\\
  \textbf{[TTT] Output:} $S^{\text{TTT}}_t$ (trained adapters).

  \begin{algorithmic}[1]
    \STATE $e_{x^{*}} = \argminF_{e_{x^{i}} \in \mathcal{Q}} \gamma(e_{x^{i}}, e_{x^{t}})$
    \IF{$\gamma(e_{x^{*}}, e_{x^{t}}) > \tau_e$}
        \STATE $S^{\text{TTT}}_t \gets \mathcal{C}_{\text{TTT}}[e_{x^{*}}]$ \COMMENT{TTT cache hit}
    \ELSE
        \STATE $S^{\text{TTT}}_t \gets \textsc{Train}(S_0, S^{\text{RAG}}_t)$ \COMMENT{Use the same retrieved samples as RAG for TTT}
        \IF{$|\mathcal{C}_{\text{TTT}}| \geq b$}
            \STATE $\mathcal{U} = \{e^{r}_{x^{1}} ... e^{r}_{x^{m}}\} = \kappa(\mathcal{C}_{\text{TTT}})$
            \STATE $\mathcal{C}_{\text{TTT}} \leftarrow \mathcal{C}_{\text{TTT}} \setminus \mathcal{U}$ 
        \ENDIF
        \STATE $\mathcal{C}_{\text{TTT}}[e_{x^{t}}] \gets S^{\text{TTT}}_t$
        \STATE $\mathcal{Q} \gets \mathcal{Q} \cup \{e_{x^{t}}\}$
    \ENDIF
  \end{algorithmic}
\end{algorithm}

\subsection{Query-State Caching (\qsc)}
\label{sec:data_reduction_strategies}

The retrieval and fine-tuning stages at the client introduce notable computational and latency overhead in the \proj pipeline. To address this, we propose Query-State Caching (\qsc), a unified caching strategy at both the retrieval (RAG) and model state (TTT) levels. As described in Algorithm~\ref{alg:qsc_unified_branches_simple} and illustrated in Figure~\ref{fig:lsm}, \qsc maintains a set of historical query embeddings $\mathcal{Q}$ and two corresponding caches: one mapping embeddings to retrieved samples for RAG, and another mapping embeddings to fine-tuned adapters for TTT. For each new query, the most similar historical embedding is identified using a similarity metric $\gamma$. If the similarity exceeds a reuse threshold $\tau_e$, the corresponding cached state (retrieved samples for RAG or adapters for TTT) is reused, allowing the system to bypass redundant retrieval or fine-tuning. Otherwise, new retrieval or fine-tuning is performed, and the cache is updated accordingly, with an eviction mechanism $\kappa$ ensuring the cache stays within a fixed budget $b$. This unified approach substantially reduces redundant computation and latency, enabling efficient and scalable test-time adaptation.

\begin{table*}[!t]
\setlength{\tabcolsep}{6pt}
\scriptsize
\centering
\renewcommand\arraystretch{1.7}
\begin{tabular}{lllc}
\toprule
\textbf{Domain} & \textbf{Task} & \textbf{Evaluation Tool} & \textbf{\# Samples}  \\
\midrule
\multirow{2}{*}{Coding} & MBPP \cite{austin2021program} & \multirow{2}{*}{Bigcode-Evaluation-Harness \cite{bigcode-evaluation-harness}} & 500  \\
& HumanEval \cite{chen2021evaluating} & & 164  \\
\cline{1-4}
\multirow{2}{*}{Math} & MathQA \cite{amini-etal-2019-mathqa} & LLM-Adapters \cite{hu2023llm_adapters} & 200\textsuperscript{*} \\
\cline{3-4}
                      & GSM-Plus \cite{li2024gsmpluscomprehensivebenchmarkevaluating} & \multirow{2}{*}{LM-Eval-Harness \citep{eval-harness}} & 200\textsuperscript{*} \\
\cline{1-2}
Medical & MedConceptsQA ATC \cite{pmlr-v174-pal22a} &  & 600\textsuperscript{*} \\
\bottomrule
\end{tabular}
\caption{
Details of the tasks evaluated in the experiments. \textsuperscript{*}Only a subset of the original test set is used for these tasks (200 samples for MathQA and GSM-Plus, 600 for ATC) to increase experimental efficiency.
}
\label{tab:evaluation_tasks}
\end{table*}

\begin{table*}[!t]
\setlength{\tabcolsep}{4pt}
\scriptsize
\centering
\renewcommand\arraystretch{1.1}
\begin{tabular}{llcccccccc}
\toprule
\textbf{Model} & \textbf{Strategy} & \textbf{MBPP} & \textbf{HumanEval} & \textbf{MathQA\textsuperscript{*}} & \textbf{GSM-Plus\textsuperscript{*}} & \textbf{ATC\textsuperscript{*}} & \textbf{Avg.} & \textbf{Impr.} & \textbf{Strategy Distribution (\%)} \\
\midrule
\multirow{5}{*}{\llamathree} 
    & No Adaptation & 51.6 & 54.9 & 29.0 & 19.5 & 36.8 & 38.4 & -- & 100\% / -- / --~\distbar{3.6}{0.1}{0.1} \\
    & RAG           & 42.0 & 48.8 & \textbf{39.5} & 20.5 & \textbf{41.8} & 38.5 & 100.4\% & -- / 100\% / --~\distbar{0.1}{3.6}{0.1} \\
    & TTT           & 49.0 & 54.9 & 37.0 & 29.5 & 36.7 & 41.4 & 108.0\% & -- / -- / 100\%~\distbar{0.1}{0.1}{3.6} \\
    & \proj         & \textbf{53.6} & 56.1 & 39.0 & 23.5 & 37.3 & 41.9 & 109.2\% & 13.3\% / 26.6\% / 60.1\%~\distbar{0.7}{1.4}{3.2} \\
    & \proj-Joint   & 50.4 & \textbf{57.9} & 35.5 & \textbf{33.0} & 39.3 & \textbf{43.2} & \textbf{112.7\%} & 13.3\% / 30.2\% / 56.6\%~\distbar{0.7}{1.6}{3.0} \\
\midrule
\multirow{5}{*}{\llamathreeone}
    & No Adaptation & 52.2 & 59.8 & 16.5 & 20.0 & 38.5 & 37.4 & -- & 100\% / -- / --~\distbar{3.6}{0.1}{0.1} \\
    & RAG           & 42.4 & 51.8 & \textbf{32.5} & 33.0 & \textbf{44.3} & 40.8 & 109.2\% & -- / 100\% / --~\distbar{0.1}{3.6}{0.1} \\
    & TTT           & 51.6 & \textbf{63.4} & 25.0 & 34.0 & 37.8 & 42.4 & 113.3\% & -- / -- / 100\%~\distbar{0.1}{0.1}{3.6} \\
    & \proj         & 54.0 & 61.0 & 29.0 & 30.5 & 38.8 & 42.7 & 114.1\% & 6.6\% / 25.8\% / 67.6\%~\distbar{0.4}{1.4}{3.6} \\
    & \proj-Joint   & \textbf{55.2} & 62.8 & 31.0 & \textbf{37.5} & 40.2 & \textbf{45.3} & \textbf{121.2\%} & 6.6\% / 23.9\% / 69.5\%~\distbar{0.2}{1.3}{3.8} \\
\midrule
\multirow{5}{*}{\mistral}
    & No Adaptation & 37.6 & 33.5 & 28.0 & 15.0 & 23.8 & 27.6 & -- & 100\% / -- / --~\distbar{3.6}{0.1}{0.1} \\
    & RAG           & 30.2 & 28.1 & \textbf{35.0} & 16.0 & \textbf{25.8} & 27.0 & 97.9\% & -- / 100\% / --~\distbar{0.1}{3.6}{0.1} \\
    & TTT           & \textbf{38.4} & \textbf{38.4} & 32.0 & 16.5 & 24.0 & \textbf{29.9} & \textbf{108.2\%} & -- / -- / 100\%~\distbar{0.1}{0.1}{3.6} \\
    & \proj         & 32.4 & 37.2 & 32.0 & 21.0 & 23.8 & 29.3 & 106.1\% & 23.2\% / 24.9\% / 51.9\%~\distbar{1.2}{1.3}{2.8} \\
    & \proj-Joint   & 33.4 & 36.6 & 32.5 & \textbf{22.5} & 24.3 & \textbf{29.9} & \textbf{108.2\%} & 23.2\% / 29.4\% / 47.4\%~\distbar{1.2}{1.5}{2.4} \\
\midrule
\multirow{5}{*}{\qwen}
    & No Adaptation & 41.2 & 26.2 & 26.0 & 32.5 & \textbf{26.5} & 30.5 & -- & 100\% / -- / --~\distbar{3.6}{0.1}{0.1} \\
    & RAG           & 38.8 & 41.5 & \textbf{32.5} & 42.0 & 26.3 & 36.2 & 118.8\% & -- / 100\% / --~\distbar{0.1}{3.6}{0.1} \\
    & TTT           & 42.4 & 25.6 & 30.5 & 37.0 & 26.3 & 32.4 & 106.2\% & -- / -- / 100\%~\distbar{0.1}{0.1}{3.6} \\
    & \proj         & 47.4 & \textbf{43.3} & 28.5 & 42.0 & \textbf{26.5} & 37.5 & 123.1\% & 42.8\% / 18.1\% / 39.1\%~\distbar{2.3}{0.9}{2.1} \\
    & \proj-Joint   & \textbf{48.0} & 42.7 & 27.5 & \textbf{43.5} & 26.2 & \textbf{37.6} & \textbf{123.2\%} & 42.8\% / 15.7\% / 41.5\%~\distbar{2.3}{0.7}{2.3} \\
\bottomrule
\end{tabular}

\caption{
Performance comparison of different adaptation strategies across representative LLMs and evaluation tasks.  ``Impr.'' denotes the relative improvement over the No Adaptation baseline. ``Strategy Distribution (\%)'' reports the proportion of queries handled by each branch in the \proj pipeline: No Adaptation, RAG, and TTT, respectively. \proj-Joint corresponds to the ``Alternative: Joint RAG and TTT Decision'' described in section~\ref{sec:system}. All reported results reflect the best performance achieved across the evaluated retrieval sample sizes $\{1, 2, 4, 8, 16\}$ for each method.
}
\label{tab:main_results_rm_06B}
\end{table*}

\begin{table*}[ht]
\centering
\small
\begin{tabular}{ll}
\toprule
\textbf{Strategy} & \textbf{Total Cost} \\
\midrule
No Adaptation & $N \cdot C_0$  \\
RAG           & $N \cdot (C_0 + C_\text{Ret} + C_\text{RAG})$ \\
TTT           & $N \cdot (C_0 + C_\text{Ret} + C_\text{TTT})$ \\
\proj         & $N \cdot (C_0 + C_\text{Rew}) + (d_\text{RAG} + d_\text{TTT}) \cdot N \cdot (C_\text{Ret} + C_\text{RAG} + C_\text{Rew}) + d_\text{TTT} \cdot N \cdot C_\text{TTT}$ \\
\proj-Joint  & $ N \cdot (C_0 + C_\text{Rew}) + (d_\text{RAG} + d_\text{TTT}) \cdot N \cdot (C_\text{Ret} + C_\text{RAG} + C_\text{TTT} + 2C_\text{Rew}) $ \\
\bottomrule
\end{tabular}
\caption{
Total cost comparison for $N$ queries under different adaptation strategies. $C_0$ denotes the base inference cost per query (No Adaptation); $C_\text{Ret}$, $C_\text{RAG}$, and $C_\text{TTT}$ represent the additional costs for retrieval, RAG, and TTT, respectively, with $C_\text{TTT} > C_\text{RAG} > 0$; $C_\text{Rew}$ is the reward model evaluation cost. For \proj, $d_\text{RAG}$ and $d_\text{TTT}$ indicate the fractions of queries routed to the RAG and TTT branches, as reported in the main results (see ``Strategy Distribution (\%)'' of Table~\ref{tab:main_results_rm_06B}).
}
\label{tab:cost_analysis}
\end{table*}

\begin{table}[htbp]
\setlength{\tabcolsep}{2pt}
\renewcommand\arraystretch{1.1}
\centering
\footnotesize
\begin{tabular}{lrrr}
\hline
\textbf{Metric} & \textbf{No Adapt.} & \textbf{RAG} & \textbf{TTT} \\
\hline
Context Length & 96.1 & 96.1 & 96.1\\
Token Generation Count & 326.6 & 353.5 & 344.7 \\
Inference Latency (sec) & 7.5 & 8.7 & 8.4 \\
\textbf{Total Latency (sec)} & \textbf{7.5} & \textbf{9.8} & \textbf{12.1} \\
\hline
\multicolumn{4}{l}{\textit{\textbf{Retrieval:}}} \\
\quad Embedding Processing (sec) & - & 0.06 & 0.06 \\
\quad Retrieval (sec) & - & 1.06 & 1.06 \\
\multicolumn{4}{l}{\textit{\textbf{RAG:}}} \\
Augmented Context Length & - & 3058.2 & -\\
\multicolumn{4}{l}{\textit{\textbf{TTT:}}} \\
\quad Train (sec) & - & - & 2.60  \\
\quad Merge (sec) & - & - & 0.01  \\
\quad Unmerge (sec) & - & - & 0.01 \\
\quad Training Token Count & - & - & 5,921.6  \\
\hline
\end{tabular}
\caption{Performance comparison of RAG and TTT against No Adaptation using \mistral on MathQA dataset. Results are averaged over 10 test samples with 8 test-time training samples per query. The experiments were conducted on NVIDIA A100.}
\label{tab:performance_analysis}
\end{table}

\begin{table*}[!t]
\setlength{\tabcolsep}{3pt}
\scriptsize
\centering
\renewcommand\arraystretch{1.1}
\begin{tabular}{lccccccccccc}
\toprule
\multirow{2}{*}{\textbf{Model}} & \textbf{Query-State} & \multirow{2}{*}{\textbf{MBPP}} & \multirow{2}{*}{\textbf{HumanEval}} & \multirow{2}{*}{\textbf{MathQA\textsuperscript{*}}} & \multirow{2}{*}{\textbf{GSM-Plus\textsuperscript{*}}} & \multirow{2}{*}{\textbf{ATC\textsuperscript{*}}} & \multirow{2}{*}{\textbf{Avg.}} & \multirow{2}{*}{\textbf{Rel.}} & \textbf{RAG Cache} & \textbf{TTT Cache} \\
 & \textbf{Caching} &  &  &  & & &  &  & \textbf{Utilization} & \textbf{Utilization} \\
\midrule
\multirow{2}{*}{\llamathree} & \ding{55} & \textbf{53.8} & 55.5 & \textbf{35.0} & \textbf{19.0} & 38.8 & \textbf{40.4} & \textbf{100.00\%} & -- & -- \\
 & \ding{51} & 52.4 & \textbf{56.1} & 33.0 & 16.5 & \textbf{39.5}& 39.5 & 97.71\% & \textbf{66.53\%} & \textbf{70.11\%} \\
\midrule
\multirow{2}{*}{\llamathreeone} & \ding{55} & \textbf{55.0} & 60.4 & \textbf{28.0} & 26.5 & 39.3 & 41.8 & 100.00\% & -- & -- \\
 & \ding{51} & 54.4 & \textbf{64.0} & 21.5 & \textbf{29.0} & \textbf{40.3} & \textbf{41.9} & \textbf{100.02\%} & \textbf{62.72\%} & \textbf{67.03\%} \\
\midrule
\multirow{2}{*}{\mistral} & \ding{55} & \textbf{41.0} & \textbf{37.2} & \textbf{30.5} & \textbf{20.5} & 23.0 & \textbf{30.4} & \textbf{100.00\%} & -- & -- \\
 & \ding{51} & 38.4 & \textbf{37.2} & 28.0 & 20.0 & \textbf{24.2} & 29.6 & 97.09\% & \textbf{62.34\%} & \textbf{70.10\%} \\
\midrule
\multirow{2}{*}{\qwen} & \ding{55} & \textbf{42.0} & 36.6 & 24.5 & \textbf{44.5} & \textbf{26.2} & \textbf{34.8} & \textbf{100.00\%} & -- & -- \\
& \ding{51} & 41.2 & \textbf{40.9} & \textbf{25.0} & 40.0 & 25.8 & 34.6 & 99.49\% & \textbf{64.02\%} & \textbf{69.85\%} \\
\bottomrule
\end{tabular}
\caption{
Performance comparison of \proj with and without Query-State Caching (\qsc). All results use 4 retrieved samples per query for fair comparison (note: this differs from Table~\ref{tab:main_results_rm_06B}, which reports the results for the best-performing retrieval sample size). ``RAG Cache Utilization'' and ``TTT Cache Utilization'' report the proportion of queries that successfully reused cached retrieval results and cached adapters, respectively, as defined in Algorithm~\ref{alg:qsc_unified_branches_simple}. These metrics reflect the effectiveness of \qsc in reducing redundant retrieval and fine-tuning operations. ``Rel.'' denotes the relative average performance compared to the baseline (\proj without \qsc).
}
\label{tab:qsc_results}
\end{table*}

\section{Experiments}
\label{sec:experiments}
We implement a prototype of the \proj system as a testbed, demonstrating the potential benefits in a larger deployment. Next, we discuss the resources utilized in our experimentation, followed by results demonstrating the benefits of enabling \proj.  
\subsection{Setup}

\paragraph{Knowledge base}

To rigorously evaluate the \proj prototype, we construct a comprehensive multi-domain knowledge base by integrating several representative datasets spanning Coding, Math, and Medical domains. This diverse collection ensures robust and generalizable evaluation across tasks. For further details regarding dataset composition and sources, please refer to Appendix \S \ref{sec:appendix_knowledge_base}.

\paragraph{Evaluation} 
We comprehensively evaluate the \proj prototype across Coding, Math, and Medical domains using a suite of established benchmarks and evaluation tools. For the coding domain, we assess performance on MBPP \cite{austin2021program} and HumanEval \cite{chen2021evaluating} using the Bigcode-Evaluation-Harness \cite{bigcode-evaluation-harness}. In the math domain, we evaluate on MathQA \cite{amini-etal-2019-mathqa} with the LLM-Adapters evaluation scripts \cite{hu2023llm_adapters}, and on GSM-Plus \cite{li2024gsmpluscomprehensivebenchmarkevaluating} using LM-Eval-Harness \citep{eval-harness}. For the medical domain, we employ MedConceptsQA ATC \cite{pmlr-v174-pal22a}, also evaluated with LM-Eval-Harness; we refer to this task as ATC for brevity. All experiments are conducted under the zero-shot setting to reflect real-world deployment scenarios.
For experimental efficiency, we evaluate \proj on subsets of some benchmarks: 200 samples for MathQA and GSM-Plus, and 600 samples for ATC. The details of all evaluation tasks are summarized in Table~\ref{tab:evaluation_tasks}.

\paragraph{Models} We test our method on various LLMs, including \llamathree, \llamathreeone \citep{dubey2024llama}, \mistral \citep{jiang2023mistral} and \qwen \citep{yang2024qwen2}. 
For the retrieval stage in \proj, we utilize \qwenemb \cite{qwen3embedding} as the embedding model.
For the reward model, we employ \skyworkzerosixb \cite{liu2025skywork}.
\paragraph{Hyperparameters and Implementation} 
For RAG and TTT, the number of retrieval samples is selected from $\{1, 2, 4, 8, 16\}$. TTT is performed for two epochs with a learning rate of $5 \times 10^{-5}$ and a batch size of 1. LoRA fine-tuning is applied in TTT, using a rank of 32 and an alpha of 16, targeting the Query, Key, Value, Up, and Down projection layers. 
The threshold $\tau_r$ is 2.0.
For \qsc, the reuse threshold $\tau_e$ and budget $b$ are set to 0.5 and 8, respectively. The similarity metric $\gamma$ is the inner product, and the eviction mechanism $\kappa$ adopts a Least Frequently Used (LFU) policy. 

\subsection{Main Results}
\label{sec:main_results}

Table~\ref{tab:main_results_rm_06B} presents a comprehensive comparison of adaptation strategies across multiple LLMs and tasks. Several key observations: (1) \proj consistently outperforms both RAG and TTT baselines, achieving the highest average accuracy improvements across all models and tasks. For example, on \llamathreeone, \proj yields a 114.1\% relative improvement over the no adaptation baseline, while the joint variant (\proj-Joint) further boosts performance to 121.2\%. (2) The joint decision strategy (\proj-Joint), which selects the best response between RAG and TTT per query, achieves the best overall results, highlighting the benefit of adaptive, reward-guided selection. (3) The strategy distribution indicates that \proj predominantly leverages TTT for challenging queries, while efficiently falling back to direct inference or RAG when appropriate, thus balancing accuracy and computational cost. (4) Notably, the effectiveness of RAG and TTT varies by task and model, underscoring the necessity of a unified, query-adaptive framework. Overall, these results demonstrate that \proj robustly enhances downstream performance across diverse domains and models, validating the effectiveness of reward-guided, collaborative test-time compute. 

\subsection{Cost Analysis} 


Table~\ref{tab:cost_analysis} presents a comparative analysis of the computational cost associated with different TTC strategies. While RAG and TTT each introduce substantial additional overhead due to retrieval, longer context or fine-tuning, the cost profile of \proj is inherently query-adaptive and cannot be strictly characterized as lower or higher than either baseline. The overall cost of \proj depends on the distribution of queries across its decision branches (see ``Strategy Distribution (\%)'' in Table~\ref{tab:main_results_rm_06B}). For queries where the initial response is sufficient, \proj terminates early, incurring only minimal inference and reward evaluation costs. However, for queries routed to RAG or TTT, \proj will incur additional overhead compared to vanilla RAG or TTT, as each branch is preceded by an initial inference and reward evaluation step. Thus, \proj embodies an adaptive early-stopping mechanism, dynamically allocating computation to maximize response quality.

To complement the theoretical cost analysis, Table~\ref{tab:performance_analysis} reports empirical measurements of computational performance across different adaptation strategies. These results, obtained on 10 test samples of MathQA, detail latency and other metrics under realistic deployment conditions. Compared to the baseline (No Adaptation), RAG and TTT increase total latency due to retrieval, augmented context and fine-tuning, with TTT incurring the highest cost. RAG significantly expands context length, while TTT introduces additional training steps; both yield higher token generation counts. 

Given that \proj might introduce additional overhead in certain scenarios, we propose Query-State Caching (QSC) to mitigate redundant retrieval and fine-tuning operations. QSC leverages historical query states to reuse previously computed results, thereby reducing unnecessary computation and latency. The effectiveness of QSC is evaluated in the following subsection.

\subsection{Query-State Caching (\qsc)} 

Table~\ref{tab:qsc_results} presents the evaluation of Query-State Caching (\qsc) across multiple LLMs and tasks. The results demonstrate that enabling \qsc yields substantial reductions in redundant retrieval and fine-tuning operations, as evidenced by high cache utilization rates for both RAG retrieval sample (62--66\%) and TTT adapter (67--70\%) caches. Importantly, \qsc achieves these efficiency gains with only marginal impact on average task performance, maintaining relative accuracy within 97--100\% of the baseline (w/o \qsc). This indicates that \qsc effectively balances efficiency and quality. 

It should be noted that the reported cache utilization rates may be somewhat optimistic due to the experimental protocol, which involves evaluating benchmark samples from the same domain in succession. This sequential testing increases the likelihood of cache hits, thereby inflating the observed utilization. In real-world deployment scenarios with more diverse and interleaved query streams, the actual cache hit rates are expected to be lower. Nevertheless, the results substantiate the potential of \qsc to reduce redundant computation while preserving robust downstream performance.

\subsection{In-Domain Sample Retrieval}
\label{sec:in-domain_retrieval}

\begin{figure}[ht]
  \centering
  \includegraphics[width=0.5\textwidth]{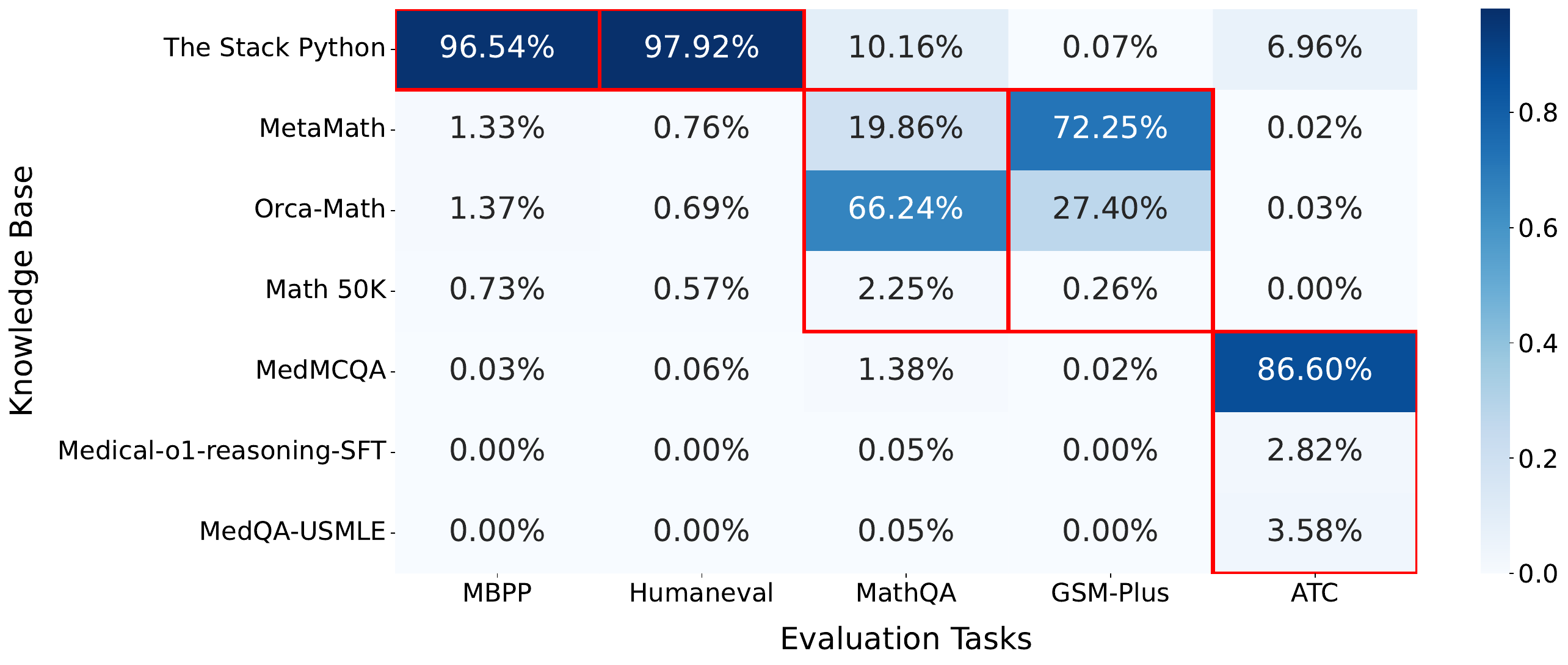}
  \caption{Domain distribution of retrieval samples from knowledge base across evaluation tasks (counted all test samples). The red border indicates the domain corresponding to the current task (in-domain). Higher values within the red border reflect stronger domain alignment.}
  \label{fig:in_domain_match_heatmap}
\end{figure}

To quantitatively assess the effectiveness of \proj in retrieving domain-relevant samples from a multi-domain knowledge base, we analyze the domain composition of retrieved samples for each evaluation task. Figure~\ref{fig:in_domain_match_heatmap} presents a heatmap of the proportional distribution of retrieved samples across all domains for five tasks.

The results demonstrate strong in-domain alignment: for each task, the majority of retrieved samples originate from the corresponding domain, as highlighted by the red-bordered cells. For instance, MBPP and Humaneval retrieve over 97\% and 98\% of samples from The Stack Python, respectively.  
This high degree of domain specificity is primarily attributable to the quality of the embedding model, which enables precise semantic matching between queries and knowledge base entries. The effectiveness of \proj in test-time adaptation fundamentally relies on both a high-quality embedding model and a well-curated, diverse knowledge base, which together ensure that retrieved samples are highly relevant to the current task.

\begin{figure}[ht]
  \centering
  \includegraphics[width=0.42\textwidth]{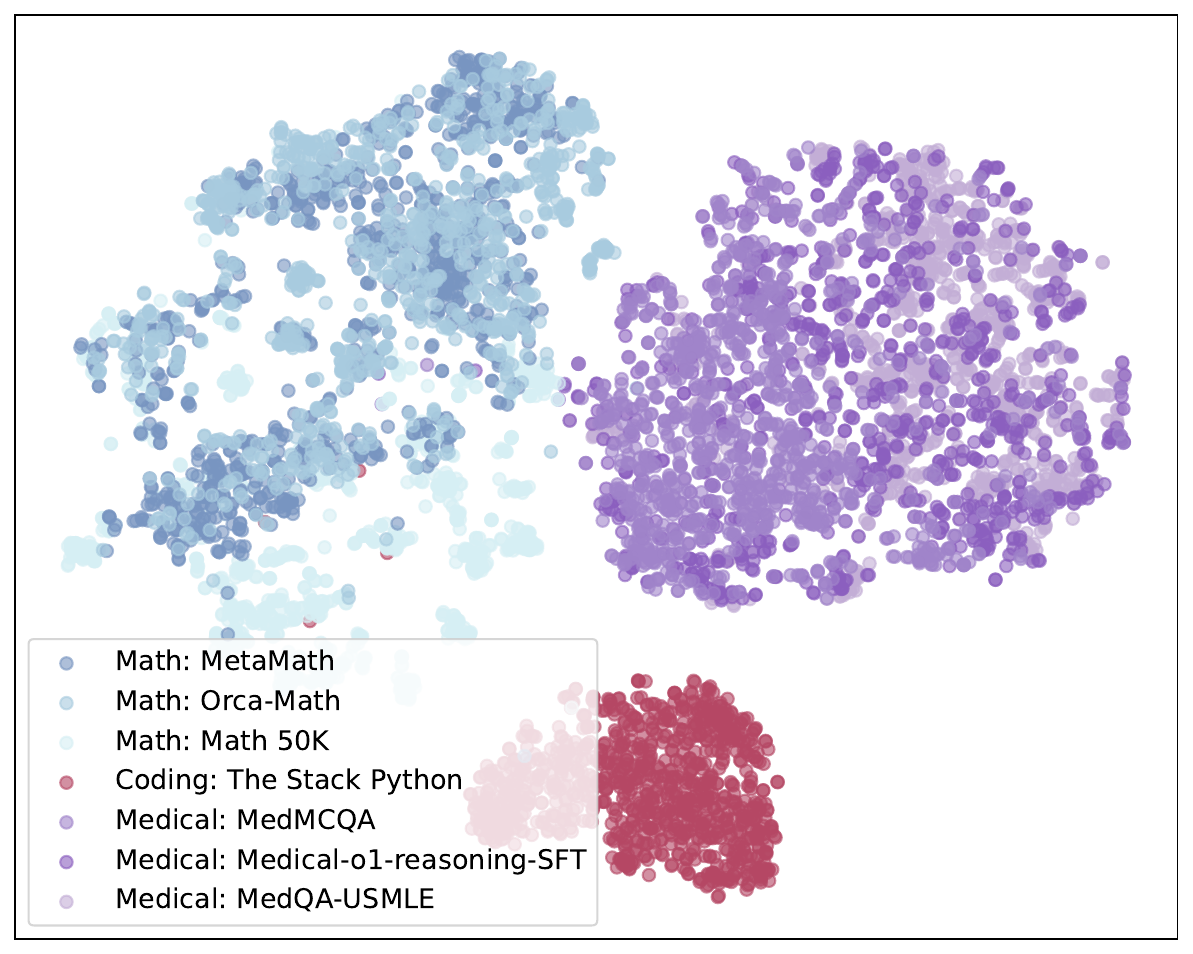}
  \caption{t-SNE visualization of the embeddings for some samples in the knowledge base using the \qwenemb embedding model. Each dataset contains 1000 randomly selected samples.}
  \label{fig:domain_tsne_visualization_qwen3}
\end{figure}

Additionally, we performed a t-SNE visualization of the embeddings for some samples in the database, as shown in Figure \ref{fig:domain_tsne_visualization_qwen3}. The visualization demonstrates a clear clustering by domain, further illustrating the discriminative power of the embedding model and the importance of database quality. These factors are essential for enabling \proj to reliably retrieve domain-specific samples, thereby maximizing the benefit of test-time adaptation.

\subsection{Extended Experimental Results}
\label{sec:extended_results}

For completeness and to facilitate deeper analysis, we provide additional experimental results and ablation studies in the Appendix. These include comprehensive comparisons across varying numbers of retrieved samples (see Appendix \S \ref{sec:appendix_retrieval_samples}, Table~\ref{tab:main_results_rm_06B_llama3_all}, Table~\ref{tab:main_results_rm_06B_llama3_1_all}, Table~\ref{tab:main_results_rm_06B_mistral_all}, and Table~\ref{tab:main_results_rm_06B_qwen_all}), reward thresholds $\tau_r$ (see Appendix \S \ref{sec:appendix_threshold} and Table~\ref{tab:threshold_analysis}), and reward models (see Appendix \S \ref{sec:appendix_rm} and Table~\ref{tab:rm_analysis}). The extended results further substantiate the robustness and generalizability of \proj across diverse models, tasks, and settings. Readers are referred to the Appendix for full tables and discussion.

\section{Conclusion}
\label{sec:conclusion}
Test-time Compute (TTC) is an effective paradigm for enhancing model performance at the expense of increased computation during inference. We present \proj, a system that adaptively selects the optimal TTC strategy for each query at test-time. The results of the \proj prototype, which utilizes supervisory signals from a knowledge base, serve as a call to action to investigate further improvements to learning at test-time methods. \proj is guided by a reward model that assists in the decision of the TTC method to apply at inference time.
Importantly, the reward-guided approach in \proj is theoretically extensible to any TTC strategy, providing an open and generalizable direction for future research in adaptive test-time compute.
The current version of \proj works on text queries. With increased sophistication, future versions of \proj must handle more complex tasks and multimodal queries.

\section*{Limitations}

Although \proj produces compelling results and demonstrates that a reward-guided approach can alleviate the challenges of deciding between popular test-time compute techniques at runtime, it also opens new research opportunities for the future. Currently, the proposed approach is limited by the manual definition of several hyperparameters, for instance determining the value for the threshold $\tau_r$ to trigger RAG or TTT. This current limitations present exciting opportunities in future work. Another current limitation is related to determining the right data mix in the knowledge base. In real-world applications, the effectiveness of \proj is likely more pronounced with a more extensive and diverse knowledge base. A larger knowledge base would provide a broader range of samples, potentially improving the relevance and quality of the data retrieved for TTT, thereby further enhancing the model's performance.
Our experimental results have demonstrated the feasibility and potential benefits of \proj.  

For experimental efficiency, our TTT experiments employ a single set of hyperparameters (learning rate, number of epochs, LoRA configuration, etc.) across all tasks and models, without extensive hyperparameter exploration. We believe that TTT has significant potential for further improvement, and that more optimal hyperparameter choices could further enhance the performance of \proj.

In the current version of \proj's prototype, the server can recover the content of the user's prompt. A real-world solution should incorporate privacy mechanisms to protect the user. In addition to encrypting the query for transmission, e.g., utilizing Secure Sockets Layer (SSL), several open research challenges exist to enhance the privacy and handling of the user's content on the server.

\section*{Ethical Considerations}

Test-time compute (TTC) techniques offer improvements in model performance, making them more accurate, albeit at the cost of increased computation. However, they alone do not solve existing challenges in large foundation models and their smaller counterparts. Our research explores systems and techniques to enable running TTC in client devices with resource constraints. However, applying our system and techniques to real-world applications must include additional safeguards to prevent hallucinations or intentional misinformation that could negatively impact the well-being of system users. The research community must continue to investigate solutions to address these and other open challenges in popular language models.

\bibliography{main}

\clearpage
\appendix
\begin{table*}[!t]
\setlength{\tabcolsep}{6pt}
\scriptsize
\centering
\renewcommand\arraystretch{1.7}
\begin{tabular}{llccl}
\toprule
\textbf{Domain} & \textbf{Dataset} & \textbf{\# Samples} & \textbf{\# Samples per Domain} & \textbf{Link}\\
\midrule
\multirow{1}{*}{Coding} & The Stack Python \cite{Kocetkov2022TheStack} & 600,000 & \multirow{1}{*}{600,000} & \href{https://huggingface.co/datasets/bigcode/the-stack}{bigcode/the-stack} \\
\cline{1-5}
\multirow{3}{*}{Math} & MetaMath \cite{yu2023metamath} & 395,000 & \multirow{3}{*}{645,035} & \href{https://huggingface.co/datasets/meta-math/MetaMathQA}{meta-math/MetaMathQA} \\
                      & Orca-Math \cite{mitra2024orcamath} & 200,035 & & \href{https://huggingface.co/datasets/microsoft/orca-math-word-problems-200k}{microsoft/orca-math-word-problems-200k} \\
                      & Math 50K \cite{hu2023llm_adapters} & 50,000 & & \href{https://github.com/AGI-Edgerunners/LLM-Adapters/blob/main/ft-training_set/math_50k.json}{math\_50k.json}\\
\cline{1-5}
\multirow{3}{*}{Medical} & MedMCQA \cite{pmlr-v174-pal22a} & 182,822 & \multirow{3}{*}{212,704} & \href{https://huggingface.co/datasets/openlifescienceai/medmcqa}{openlifescienceai/medmcqa} \\
                         & Medical-o1-reasoning-SFT \cite{chen2024huatuogpto1medicalcomplexreasoning} & 19,704 & & \href{https://huggingface.co/datasets/FreedomIntelligence/medical-o1-reasoning-SFT}{FreedomIntelligence/medical-o1-reasoning-SFT}\\
                         & MedQA-USMLE \cite{jin2020disease} & 10,178 & & \href{https://huggingface.co/datasets/GBaker/MedQA-USMLE-4-options-hf}{GBaker/MedQA-USMLE-4-options-hf}\\
\cline{1-5}
 & & \textbf{Total:} & \textbf{1,457,739} \\
\bottomrule
\end{tabular}
\caption{
Knowledge base composition. These datasets cover various domains, including coding, math, and medical. The Stack Python dataset consists of a random sample of 600,000 entries from the original Stack dataset (Python). 
}
\label{tab:knowledge_base}
\end{table*}

\section{Knowledge Base Details}
\label{sec:appendix_knowledge_base}

To support rigorous and multi-domain evaluation of the \proj system, we aggregate a large-scale knowledge base comprising datasets from Coding, Math, and Medical domains. Table~\ref{tab:knowledge_base} summarizes the dataset composition, sample counts, and sources.

\paragraph{Coding Domain}
We sample 600,000 entries from the Python subset of The Stack~\cite{Kocetkov2022TheStack}, providing diverse programming knowledge.

\paragraph{Math Domain}
Three datasets are included: MetaMath~\cite{yu2023metamath} (395,000 samples), Orca-Math~\cite{mitra2024orcamath} (200,035 samples), and Math 50K~\cite{hu2023llm_adapters} (50,000 samples), collectively covering a wide range of mathematical reasoning and problem-solving scenarios.

\paragraph{Medical Domain}
The medical subset consists of MedMCQA~\cite{pmlr-v174-pal22a} (182,822 samples), Medical-o1-reasoning-SFT~\cite{chen2024huatuogpto1medicalcomplexreasoning} (19,704 samples), and MedQA-USMLE~\cite{jin2020disease} (10,178 samples), focusing on medical question answering and reasoning.

All datasets are strictly used for experimental purposes. This knowledge base provides a robust foundation for cross-domain adaptation and benchmarking.

\begin{table*}[!t]
\setlength{\tabcolsep}{8pt}
\scriptsize
\centering
\renewcommand\arraystretch{1.1}
\begin{tabular}{llccccccc}
\toprule
\textbf{Number of Retrieval Samples} & \textbf{Strategy} & \textbf{MBPP} & \textbf{HumanEval} & \textbf{MathQA\textsuperscript{*}} & \textbf{GSM-Plus\textsuperscript{*}} & \textbf{ATC\textsuperscript{*}} & \textbf{Avg.} & \textbf{Impr.} \\
\midrule

 /   & No Adaptation & 51.6 & 54.9 & 29.0 & 19.5 & 36.8 & 38.4 & -- \\
\midrule
\multirow{4}{*}{1}
    & RAG           & 36.6 & 41.5 & 36.5 & 25.5 & 38.2 & 35.7 & 92.9\% \\
    & TTT           & 52.2 & 55.5 & 28.5 & 20.5 & 36.8 & 38.7 & 100.9\% \\
    & \proj         & 52.8 & 53.7 & 32.0 & 27.5 & 37.3 & 40.7 & 106.0\% \\
    & \proj-Joint   & 53.4 & 54.9 & 33.5 & 30.0 & 37.5 & \textbf{41.9} & \textbf{109.1\%} \\
\midrule
\multirow{4}{*}{2}
    & RAG           & 40.0 & 45.7 & 37.0 & 28.0 & 40.3 & 38.2 & 99.6\% \\
    & TTT           & 51.8 & 56.1 & 27.0 & 21.5 & 37.2 & 38.7 & 100.9\% \\
    & \proj         & 52.0 & 54.9 & 30.5 & 27.0 & 37.5 & 40.4 & 105.3\% \\
    & \proj-Joint   & 52.6 & 54.9 & 31.0 & 26.0 & 38.5 & \textbf{40.6} & \textbf{105.8\%} \\
\midrule
\multirow{4}{*}{4}
    & RAG           & 42.0 & 48.8 & 39.5 & 20.5 & 41.8 & 38.5 & 100.4\% \\
    & TTT           & 52.2 & 56.1 & 35.5 & 19.5 & 37.2 & 40.1 & 104.5\% \\
    & \proj         & 53.6 & 56.1 & 39.0 & 23.5 & 37.3 & 41.9 & 109.2\% \\
    & \proj-Joint   & 54.8 & 56.7 & 41.5 & 21.5 & 37.2 & \textbf{42.3} & \textbf{110.4\%} \\
\midrule
\multirow{4}{*}{8}
    & RAG           & 42.6 & 49.4 & 35.5 & 14.0 & 41.8 & 36.7 & 95.6\% \\
    & TTT           & 52.0 & 57.9 & 40.5 & 16.0 & 37.5 & \textbf{40.8} & \textbf{106.3\%} \\
    & \proj         & 51.8 & 54.9 & 34.0 & 19.0 & 36.0 & 39.1 & 102.0\% \\
    & \proj-Joint   & 52.4 & 56.1 & 37.0 & 18.5 & 38.5 & 40.5 & 105.6\% \\
\midrule
\multirow{4}{*}{16}
    & RAG           & 39.6 & 48.2 & 37.0 & 11.5 & 41.5 & 35.6 & 92.7\% \\
    & TTT           & 49.0 & 54.9 & 37.0 & 29.5 & 36.7 & 41.4 & 108.0\% \\
    & \proj         & 49.6 & 54.3 & 32.5 & 24.0 & 36.8 & 39.4 & 102.8\% \\
    & \proj-Joint   & 50.4 & 57.9 & 35.5 & 33.0 & 39.3 & \textbf{43.2} & \textbf{112.7\%} \\
\bottomrule
\end{tabular}

\caption{
Performance comparison of different adaptation strategies for \llamathree. ``Impr.'' denotes the relative improvement over the No Adaptation baseline. \proj-Joint corresponds to the ``Alternative: Joint RAG and TTT Decision'' described in section~\ref{sec:system}.
}
\label{tab:main_results_rm_06B_llama3_all}
\end{table*}

\begin{table*}[!t]
\setlength{\tabcolsep}{8pt}
\scriptsize
\centering
\renewcommand\arraystretch{1.1}
\begin{tabular}{llccccccc}
\toprule
\textbf{Number of Retrieval Samples} & \textbf{Strategy} & \textbf{MBPP} & \textbf{HumanEval} & \textbf{MathQA\textsuperscript{*}} & \textbf{GSM-Plus\textsuperscript{*}} & \textbf{ATC\textsuperscript{*}} & \textbf{Avg.} & \textbf{Impr.} \\
\midrule

  /   & No Adaptation & 52.2 & 59.8 & 16.5 & 20.0 & 38.5 & 37.4 & -- \\
\midrule
\multirow{4}{*}{1}
    & RAG           & 38.4 & 45.7 & 26.0 & 41.5 & 41.5 & 38.6 & 103.3\% \\
    & TTT           & 51.4 & 60.4 & 18.0 & 18.5 & 38.8 & 37.4 & 100.1\% \\
    & \proj         & 52.6 & 60.4 & 21.5 & 31.0 & 38.8 & \textbf{40.9} & \textbf{109.3\%} \\
    & \proj-Joint   & 52.8 & 59.8 & 21.5 & 30.0 & 39.7 & 40.8 & 109.0\% \\
\midrule
\multirow{4}{*}{2}
    & RAG           & 42.4 & 51.8 & 32.5 & 33.0 & 44.3 & 40.8 & 109.2\% \\
    & TTT           & 51.4 & 60.4 & 19.0 & 23.0 & 39.0 & 38.6 & 103.1\% \\
    & \proj         & 51.6 & 59.2 & 24.0 & 29.0 & 41.3 & 41.0 & 109.7\% \\
    & \proj-Joint   & 52.6 & 59.2 & 25.0 & 30.0 & 40.8 & \textbf{41.5} & \textbf{111.0\%} \\
\midrule
\multirow{4}{*}{4}
    & RAG           & 45.0 & 50.6 & 28.0 & 36.0 & 43.5 & 40.6 & 108.6\% \\
    & TTT           & 52.2 & 61.0 & 17.0 & 20.0 & 38.8 & 37.8 & 101.1\% \\
    & \proj         & 55.2 & 61.0 & 22.0 & 25.0 & 40.0 & 40.6 & 108.7\% \\
    & \proj-Joint   & 54.8 & 60.4 & 22.5 & 30.0 & 40.3 & \textbf{41.6} & \textbf{111.3\%} \\
\midrule
\multirow{4}{*}{8}
    & RAG           & 41.8 & 53.1 & 32.5 & 23.0 & 43.5 & 38.8 & 103.7\% \\
    & TTT           & 51.6 & 61.0 & 20.0 & 22.0 & 39.5 & 38.8 & 103.8\% \\
    & \proj         & 53.0 & 59.2 & 26.5 & 25.5 & 41.0 & 41.0 & 109.7\% \\
    & \proj-Joint   & 53.0 & 60.4 & 27.5 & 27.5 & 41.5 & \textbf{42.0} & \textbf{112.3\%} \\
\midrule
\multirow{4}{*}{16}
    & RAG           & 38.0 & 54.9 & 31.0 & 21.0 & 41.2 & 37.2 & 99.5\% \\
    & TTT           & 51.6 & 63.4 & 25.0 & 34.0 & 37.8 & 42.4 & 113.3\% \\
    & \proj         & 54.0 & 61.0 & 29.0 & 30.5 & 38.8 & 42.7 & 114.1\% \\
    & \proj-Joint   & 55.2 & 62.8 & 31.0 & 37.5 & 40.2 & \textbf{45.3} & \textbf{121.2\%} \\
\bottomrule
\end{tabular}

\caption{
Performance comparison of different adaptation strategies for \llamathreeone. ``Impr.'' denotes the relative improvement over the No Adaptation baseline. \proj-Joint corresponds to the ``Alternative: Joint RAG and TTT Decision'' described in section~\ref{sec:system}.
}
\label{tab:main_results_rm_06B_llama3_1_all}
\end{table*}

\begin{table*}[!t]
\setlength{\tabcolsep}{8pt}
\scriptsize
\centering
\renewcommand\arraystretch{1.1}
\begin{tabular}{llccccccc}
\toprule
\textbf{Number of Retrieval Samples} & \textbf{Strategy} & \textbf{MBPP} & \textbf{HumanEval} & \textbf{MathQA\textsuperscript{*}} & \textbf{GSM-Plus\textsuperscript{*}} & \textbf{ATC\textsuperscript{*}} & \textbf{Avg.} & \textbf{Impr.} \\
\midrule

/   & No Adaptation & 37.6 & 33.5 & 28.0 & 15.0 & 23.8 & 27.6 & -- \\
\midrule
\multirow{4}{*}{1}
    & RAG           & 24.6 & 26.8 & 29.5 & 25.0 & 26.5 & 26.5 & 96.0\% \\
    & TTT           & 38.4 & 38.4 & 32.0 & 16.5 & 24.0 & \textbf{29.9} & \textbf{108.2\%} \\
    & \proj         & 29.0 & 34.8 & 32.0 & 22.0 & 26.2 & 28.8 & 104.3\% \\
    & \proj-Joint   & 29.6 & 37.2 & 32.0 & 23.0 & 24.8 & 29.3 & 106.3\% \\
\midrule
\multirow{4}{*}{2}
    & RAG           & 25.4 & 23.8 & 33.0 & 22.0 & 25.0 & 25.8 & 93.6\% \\
    & TTT           & 40.2 & 36.6 & 29.0 & 14.0 & 23.5 & 28.7 & 103.9\% \\
    & \proj         & 31.2 & 37.8 & 34.0 & 20.0 & 23.0 & 29.2 & 105.8\% \\
    & \proj-Joint   & 32.4 & 36.6 & 36.5 & 20.5 & 22.7 & \textbf{29.7} & \textbf{107.8\%} \\
\midrule
\multirow{4}{*}{4}
    & RAG           & 27.8 & 27.4 & 29.5 & 16.0 & 24.7 & 25.1 & 90.9\% \\
    & TTT           & 39.0 & 37.8 & 30.5 & 15.0 & 23.3 & \textbf{29.1} & \textbf{105.6\%} \\
    & \proj         & 30.6 & 37.2 & 32.0 & 17.5 & 23.3 & 28.1 & 101.9\% \\
    & \proj-Joint   & 31.8 & 36.6 & 32.5 & 18.0 & 24.5 & 28.7 & 103.9\% \\
\midrule
\multirow{4}{*}{8}
    & RAG           & 29.4 & 22.0 & 32.5 & 19.5 & 26.3 & 25.9 & 94.0\% \\
    & TTT           & 39.8 & 32.3 & 31.5 & 19.0 & 24.0 & 29.3 & 106.3\% \\
    & \proj         & 32.4 & 37.2 & 32.0 & 21.0 & 23.8 & 29.3 & 106.1\% \\
    & \proj-Joint   & 33.4 & 36.6 & 32.5 & 22.5 & 24.3 & \textbf{29.9} & \textbf{108.2\%} \\
\midrule
\multirow{4}{*}{16}
    & RAG           & 30.2 & 28.1 & 35.0 & 16.0 & 25.8 & 27.0 & 97.9\% \\
    & TTT           & 36.4 & 29.9 & 30.0 & 13.0 & 24.8 & 26.8 & 97.2\% \\
    & \proj         & 30.8 & 34.2 & 35.5 & 19.0 & 23.8 & 28.7 & 103.9\% \\
    & \proj-Joint   & 31.4 & 36.6 & 35.5 & 19.5 & 24.7 & \textbf{29.5} & \textbf{107.0\%} \\
\bottomrule
\end{tabular}

\caption{
Performance comparison of different adaptation strategies for \mistral. ``Impr.'' denotes the relative improvement over the No Adaptation baseline. \proj-Joint corresponds to the ``Alternative: Joint RAG and TTT Decision'' described in section~\ref{sec:system}. 
}
\label{tab:main_results_rm_06B_mistral_all}
\end{table*}

\begin{table*}[!t]
\setlength{\tabcolsep}{8pt}
\scriptsize
\centering
\renewcommand\arraystretch{1.1}
\begin{tabular}{llccccccc}
\toprule
\textbf{Number of Retrieval Samples} & \textbf{Strategy} & \textbf{MBPP} & \textbf{HumanEval} & \textbf{MathQA\textsuperscript{*}} & \textbf{GSM-Plus\textsuperscript{*}} & \textbf{ATC\textsuperscript{*}} & \textbf{Avg.} & \textbf{Impr.} \\
\midrule

/   & No Adaptation & 41.2 & 26.2 & 26.0 & 32.5 & 26.5 & 30.5 & -- \\
\midrule
\multirow{4}{*}{1}
    & RAG           & 29.2 & 23.8 & 27.5 & 30.0 & 26.8 & 27.5 & 90.1\% \\
    & TTT           & 42.4 & 27.4 & 23.5 & 32.0 & 26.0 & 30.3 & 99.3\% \\
    & \proj         & 45.4 & 40.2 & 25.5 & 36.0 & 26.3 & 34.7 & 113.8\% \\
    & \proj-Joint   & 45.4 & 41.5 & 26.0 & 36.5 & 25.5 & \textbf{35.0} & \textbf{114.7\%} \\
\midrule
\multirow{4}{*}{2}
    & RAG           & 32.6 & 25.6 & 32.0 & 36.0 & 25.5 & 30.3 & 99.5\% \\
    & TTT           & 39.8 & 26.8 & 24.0 & 31.0 & 25.8 & 29.5 & 96.8\% \\
    & \proj         & 45.4 & 39.0 & 27.5 & 41.0 & 25.2 & 35.6 & 116.8\% \\
    & \proj-Joint   & 45.2 & 40.2 & 27.0 & 41.5 & 26.0 & \textbf{36.0} & \textbf{118.1\%} \\
\midrule
\multirow{4}{*}{4}
    & RAG           & 36.6 & 36.0 & 30.5 & 40.5 & 27.0 & 34.1 & 111.9\% \\
    & TTT           & 41.2 & 27.4 & 22.5 & 32.5 & 26.0 & 29.9 & 98.2\% \\
    & \proj         & 46.0 & 41.5 & 26.5 & 41.5 & 26.2 & 36.3 & 119.2\% \\
    & \proj-Joint   & 45.6 & 43.9 & 27.0 & 43.5 & 26.0 & \textbf{37.2} & \textbf{122.0\%} \\
\midrule
\multirow{4}{*}{8}
    & RAG           & 38.8 & 41.5 & 32.5 & 42.0 & 26.3 & 36.2 & 118.8\% \\
    & TTT           & 41.8 & 27.4 & 26.5 & 35.0 & 26.2 & 31.4 & 103.0\% \\
    & \proj         & 47.4 & 43.3 & 28.5 & 42.0 & 26.5 & 37.5 & 123.1\% \\
    & \proj-Joint   & 48.0 & 42.7 & 27.5 & 43.5 & 26.2 & \textbf{37.6} & \textbf{123.2\%} \\
\midrule
\multirow{4}{*}{16}
    & RAG           & 35.4 & 37.8 & 32.5 & 46.0 & 25.7 & 35.5 & 116.4\% \\
    & TTT           & 42.4 & 25.6 & 30.5 & 37.0 & 26.3 & 32.4 & 106.2\% \\
    & \proj         & 46.2 & 41.5 & 28.0 & 45.5 & 25.7 & 37.4 & 122.6\% \\
    & \proj-Joint   & 46.4 & 45.1 & 29.0 & 46.0 & 26.3 & \textbf{38.6} & \textbf{126.5\%} \\
\bottomrule
\end{tabular}

\caption{
Performance comparison of different adaptation strategies for \qwen. ``Impr.'' denotes the relative improvement over the No Adaptation baseline. \proj-Joint corresponds to the ``Alternative: Joint RAG and TTT Decision'' described in section~\ref{sec:system}. 
}
\label{tab:main_results_rm_06B_qwen_all}
\end{table*}

\section{Detailed Comparison Across Retrieval Sample Sizes}
\label{sec:appendix_retrieval_samples}

Tables~\ref{tab:main_results_rm_06B_llama3_all}--\ref{tab:main_results_rm_06B_qwen_all} present comprehensive results for all evaluated models and adaptation strategies under varying numbers of retrieved samples. Across all settings, both \proj and \proj-Joint consistently outperform baseline RAG and TTT approaches, achieving the highest average accuracy and relative improvement. This superiority holds regardless of the retrieval sample size, demonstrating the robustness and effectiveness of reward-guided, query-adaptive selection in collaborative test-time compute.

\begin{table*}[!t]
\setlength{\tabcolsep}{2pt}
\scriptsize
\centering
\renewcommand\arraystretch{1.5}
\begin{tabular}{llccccccccc}
\toprule
\textbf{Model} & \textbf{Strategy} & \textbf{Threshold ($\tau_r$)} & \textbf{MBPP} & \textbf{HumanEval} & \textbf{MathQA\textsuperscript{*}} & \textbf{GSM-Plus\textsuperscript{*}} & \textbf{ATC\textsuperscript{*}} & \textbf{Avg.} & \textbf{Impr.} & \textbf{Strategy Distribution (\%)} \\
\midrule
\multirow{9}{*}{\llamathree} 
    & No Adaptation & - & 51.6 & 54.9 & 29.0 & 19.5 & 36.8 & 38.4 & -- & 100\% / -- / -- \\
    & RAG           & - & 42.0 & 48.8 & 39.5 & 20.5 & 41.8 & 38.5 & 100.4\% & -- / 100\% / --\\
    & TTT           & - & 49.0 & 54.9 & 37.0 & 29.5 & 36.7 & 41.4 & 108.0\% & -- / -- / 100\% \\
    \cdashline{2-11}
    & \multirow{3}{*}{\proj}   & 2.0 & 53.6 & 56.1 & 39.0 & 23.5 & 37.3 & 41.9 & 109.2\% & 13.3\% / 26.6\% / 60.1\% \\
    &     & 5.0 & 53.4 & 56.1 & 39.5 & 22.0 & 37.3 & 41.7 & 108.6\% & 1.2\% / 26.9\% / 71.9\% \\
    &     & 8.0 & 53.4 & 56.1 & 39.5 & 22.0 & 36.8 & 41.6 & 108.4\% & 0.1\% / 27.0\% / 73.0\% \\
    \cdashline{2-11}
    & \multirow{3}{*}{\proj-Joint}  & 2.0 & 50.4 & 57.9 & 35.5 & 33.0 & 39.3 & 43.2 & 112.7\% & 13.3\% / 30.2\% / 56.6\%\\
    &     & 5.0 & 54.6 & 56.7 & 42.0 & 20.5 & 37.7 & 42.3 & 110.3\% & 1.2\% / 28.2\% / 70.6\% \\
    &     & 8.0 & 54.6 & 56.7 & 42.0 & 20.5 & 38.2 & 42.4 & 110.5\% & 0.1\% / 28.4\% / 71.6\% \\
\midrule
\multirow{9}{*}{\llamathreeone}
    & No Adaptation & - & 52.2 & 59.8 & 16.5 & 20.0 & 38.5 & 37.4 & -- & 100\% / -- / -- \\
    & RAG           & - & 42.4 & 51.8 & 32.5 & 33.0 & 44.3 & 40.8 & 109.2\% & -- / 100\% / -- \\
    & TTT           & - & 51.6 & 63.4 & 25.0 & 34.0 & 37.8 & 42.4 & 113.3\% & -- / -- / 100\%\\
    \cdashline{2-11}
    & \multirow{3}{*}{\proj}   & 2.0      & 54.0 & 61.0 & 29.0 & 30.5 & 38.8 & 42.7 & 114.1\% & 6.6\% / 25.8\% / 67.6\% \\
    &     & 5.0 & 53.6 & 62.2 & 30.5 & 30.0 & 38.8 & 43.0 & 115.1\% & 1.7\% / 23.3\% / 75.0\% \\
    &     & 8.0 & 53.6 & 62.2 & 30.5 & 29.5 & 38.8 & 42.9 & 114.8\% & 0.0\% / 23.3\% / 76.7\% \\
    \cdashline{2-11}
    & \multirow{3}{*}{\proj-Joint} & 2.0  & 55.2 & 62.8 & 31.0 & 37.5 & 40.2 & 45.3 & 121.2\% & 6.6\% / 23.9\% / 69.5\%\\
    &     & 5.0 & 55.2 & 63.4 & 32.5 & 37.0 & 39.8 & 45.6 & 121.9\% & 1.7\% / 24.4\% / 73.9\% \\
    &     & 8.0 & 55.2 & 64.0 & 32.5 & 37.0 & 40.2 & 45.8 & 122.4\% & 0.0\% / 24.6\% / 75.4\% \\
\midrule
\multirow{9}{*}{\mistral}
    & No Adaptation & - & 37.6 & 33.5 & 28.0 & 15.0 & 23.8 & 27.6 & -- & 100\% / -- / --\\
    & RAG           & - & 30.2 & 28.1 & 35.0 & 16.0 & 25.8 & 27.0 & 97.9\% & -- / 100\% / -- \\
    & TTT           & - & 38.4 & 38.4 & 32.0 & 16.5 & 24.0 & 29.9 & 108.2\% & -- / -- / 100\%\\
    \cdashline{2-11}
    & \multirow{3}{*}{\proj}    & 2.0     & 32.4 & 37.2 & 32.0 & 21.0 & 23.8 & 29.3 & 106.1\% & 23.2\% / 24.9\% / 51.9\%\\
    &     & 5.0 & 32.4 & 37.8 & 34.5 & 21.0 & 23.2 & 29.8 & 107.9\% & 4.6\% / 32.2\% / 63.2\%\\
    &     & 8.0 & 32.8 & 37.8 & 37.5 & 19.5 & 23.3 & 30.2 & 109.4\% & 1.4\% / 33.1\% / 65.5\% \\
    \cdashline{2-11}
    & \multirow{3}{*}{\proj-Joint} & 2.0 & 33.4 & 36.6 & 32.5 & 22.5 & 24.3 & 29.9 & 108.2\% & 23.2\% / 29.4\% / 47.4\% \\
    &     & 5.0 & 35.2 & 36.0 & 31.5 & 24.0 & 24.3 & 30.2 & 109.5\% & 4.6\% / 35.0\% / 60.4\% \\
    &     & 8.0 & 33.8 & 36.6 & 39.0 & 20.5 & 22.7 & 30.5 & 110.6\% & 1.4\% / 32.9\% / 65.8\% \\
\midrule
\multirow{9}{*}{\qwen}
    & No Adaptation & - & 41.2 & 26.2 & 26.0 & 32.5 & 26.5 & 30.5 & -- & 100\% / -- / --\\
    & RAG           & - & 38.8 & 41.5 & 32.5 & 42.0 & 26.3 & 36.2 & 118.8\% & -- / 100\% / -- \\
    & TTT           & - & 42.4 & 25.6 & 30.5 & 37.0 & 26.3 & 32.4 & 106.2\% & -- / -- / 100\% \\
    \cdashline{2-11}
    & \multirow{3}{*}{\proj}  & 2.0       & 47.4 & 43.3 & 28.5 & 42.0 & 26.5 & 37.5 & 123.1\% & 42.8\% / 18.1\% / 39.1\%\\
    &     & 5.0 & 46.6 & 42.7 & 31.5 & 43.5 & 27.2 & 38.3 & 125.6\% & 12.1\% / 24.0\% / 63.9\% \\
    &     & 8.0 & 46.0 & 39.0 & 31.0 & 50.0 & 25.0 & 38.2 & 125.3\% & 4.5\% / 25.4\% / 70.1\% \\
    \cdashline{2-11}
    & \multirow{3}{*}{\proj-Joint}  & 2.0 & 48.0 & 42.7 & 27.5 & 43.5 & 26.2 & 37.6 & 123.2\% & 42.8\% / 15.7\% / 41.5\%\\
    &     & 5.0 & 46.4 & 45.1 & 31.0 & 49.5 & 26.0 & 39.6 & 129.9\% & 12.1\% / 21.6\% / 66.4\% \\
    &     & 8.0 & 46.4 & 43.9 & 32.5 & 51.5 & 26.2 & 40.1 & 131.5\% & 4.5\% / 24.7\% / 70.8\% \\
\bottomrule
\end{tabular}

\caption{
\textbf{Impact of Reward Threshold in \proj.}
Performance comparison of different adaptation strategies under varying reward thresholds. The ``Threshold ($\tau_r$)'' column denotes the value of the reward model's first-stage evaluation parameter: a higher threshold enforces stricter quality requirements for the initial response, increasing the likelihood of triggering RAG or TTT adaptation. Larger thresholds generally yield higher accuracy, but also incur greater computational cost due to more frequent execution of advanced adaptation strategies. ``Impr.'' denotes the relative improvement over the No Adaptation baseline. ``Strategy Distribution (\%)'' reports the proportion of queries handled by each branch in the \proj pipeline: No Adaptation, RAG, and TTT, respectively.
}
\label{tab:threshold_analysis}
\end{table*}

\section{Analysis of Reward Threshold}
\label{sec:appendix_threshold}

Table~\ref{tab:threshold_analysis} presents an evaluation of the effect of the reward threshold (hyper-parameter) on the selection and effectiveness of adaptation strategies within the \proj framework. The threshold controls the minimum quality required for the initial model response, as assessed by the reward model. Queries with reward scores below this threshold are routed to more advanced adaptation stages (RAG or TTT).
\paragraph{Accuracy vs. Cost Trade-off:} As the threshold increases, the average accuracy and relative improvement often improve, reflecting the benefit of more advanced adaptation. However, this comes at the expense of increased computational cost, since more queries undergo RAG and/or fine-tuning. The distribution of queries across No Adaptation, RAG, and TTT branches shifts toward greater use of adaptation strategies as the threshold rises, further illustrating the trade-off between performance and efficiency.

In summary, the reward threshold is a critical parameter for balancing accuracy and efficiency in adaptive test-time compute. Careful tuning is required to achieve optimal results for specific deployment scenarios.

\begin{table*}[!t]
\setlength{\tabcolsep}{1.5pt}
\scriptsize
\centering
\renewcommand\arraystretch{1.5}
\begin{tabular}{llccccccccccc}
\toprule
\multirow{2}{*}{\textbf{Model}} & \multirow{2}{*}{\textbf{Strategy}} & \textbf{Threshold} & \textbf{Reward} & \multirow{2}{*}{\textbf{MBPP}} & \multirow{2}{*}{\textbf{HumanEval}} & \multirow{2}{*}{\textbf{MathQA\textsuperscript{*}}} & \multirow{2}{*}{\textbf{GSM-Plus\textsuperscript{*}}} & \multirow{2}{*}{\textbf{ATC\textsuperscript{*}}} & \multirow{2}{*}{\textbf{Avg.}} & \multirow{2}{*}{\textbf{Impr.}} & \textbf{Strategy} \\
 &  &  \textbf{($\tau_r$)} & \textbf{Model Size} & & &  &  & &  & & \textbf{Distribution (\%)} \\
\midrule
\multirow{9}{*}{\llamathree} 
    & No Adaptation & - & - & 51.6 & 54.9 & 29.0 & 19.5 & 36.8 & 38.4 & -- & 100\% / -- / -- \\
    & RAG           & - & - & 42.0 & 48.8 & 39.5 & 20.5 & 41.8 & 38.5 & 100.4\% & -- / 100\% / --\\
    & TTT           & - & - & 49.0 & 54.9 & 37.0 & 29.5 & 36.7 & 41.4 & 108.0\% & -- / -- / 100\% \\
    \cdashline{2-12}
    &  \multirow{4}{*}{\proj}   & \multirow{2}{*}{5.0} & 0.6B & 53.4 & 56.1 & 39.5 & 22.0 & 37.3 & 41.7 & 108.6\% & 1.2\% / 26.9\% / 71.9\% \\
    &   & & 8B & 53.6 & 54.9 & 34.0 & 28.5 & 38.8 & 42.0 & 109.4\% & 38.5\% / 25.4\% / 36.1\% \\
    \cdashline{3-12}
    &     & \multirow{2}{*}{8.0} & 0.6B & 53.4 & 56.1 & 39.5 & 22.0 & 36.8 & 41.6 & 108.4\% & 0.1\% / 27.0\% / 73.0\% \\
    &   & & 8B & 53.4 & 55.5 & 33.5 & 29.5 & 37.5 & 41.9 & 109.2\% & 19.6\% / 28.2\% / 52.2\% \\
    \cdashline{2-12}
    &  \multirow{4}{*}{\proj-Joint}   & \multirow{2}{*}{5.0} & 0.6B & 54.6 & 56.7 & 42.0 & 20.5 & 37.7 & 42.3 & 110.3\% & 1.2\% / 28.2\% / 70.6\% \\
    &   & & 8B & 52.0 & 57.3 & 34.0 & 31.0 & 37.7 & 42.4 & 110.5\% & 38.5\% / 22.5\% / 39.0\% \\
    \cdashline{3-12}
    &     & \multirow{2}{*}{8.0} & 0.6B & 54.6 & 56.7 & 42.0 & 20.5 & 38.2 & 42.4 & 110.5\% & 0.1\% / 28.4\% / 71.6\% \\
    &   & & 8B & 53.8 & 56.1 & 32.5 & 30.5 & 38.0 & 42.2 & 110.0\% & 19.6\% / 28.5\% / 51.9\% \\
\midrule
\multirow{9}{*}{\llamathreeone}
    & No Adaptation & - & - & 52.2 & 59.8 & 16.5 & 20.0 & 38.5 & 37.4 & -- & 100\% / -- / -- \\
    & RAG           & - & - & 42.4 & 51.8 & 32.5 & 33.0 & 44.3 & 40.8 & 109.2\% & -- / 100\% / -- \\
    & TTT           & - & - & 51.6 & 63.4 & 25.0 & 34.0 & 37.8 & 42.4 & 113.3\% & -- / -- / 100\%\\
    \cdashline{2-12}
    &  \multirow{4}{*}{\proj}   & \multirow{2}{*}{5.0} & 0.6B & 53.6 & 62.2 & 30.5 & 30.0 & 38.8 & 43.0 & 115.1\% & 1.7\% / 23.3\% / 75.0\% \\
    &   & & 8B & 53.6 & 59.8 & 21.5 & 31.0 & 41.2 & 41.4 & 110.7\% & 45.1\% / 19.0\% / 35.9\% \\
    \cdashline{3-12}
    &     & \multirow{2}{*}{8.0} & 0.6B & 53.6 & 62.2 & 30.5 & 29.5 & 38.8 & 42.9 & 114.8\% & 0.0\% / 23.3\% / 76.7\% \\
    &   & & 8B & 53.8 & 59.8 & 27.5 & 31.0 & 39.3 & 42.3 & 113.1\% & 27.3\% / 18.2\% / 54.6\% \\
    \cdashline{2-12}
    &  \multirow{4}{*}{\proj-Joint}   & \multirow{2}{*}{5.0} & 0.6B & 55.2 & 63.4 & 32.5 & 37.0 & 39.8 & 45.6 & 121.9\% & 1.7\% / 24.4\% / 73.9\% \\
    &   & & 8B & 53.8 & 59.8 & 24.5 & 36.5 & 40.0 & 42.9 & 114.8\% & 45.1\% / 13.5\% / 41.4\% \\
    \cdashline{3-12}
    &     & \multirow{2}{*}{8.0} & 0.6B & 55.2 & 64.0 & 32.5 & 37.0 & 40.2 & 45.8 & 122.4\% & 0.0\% / 24.6\% / 75.4\% \\
    &   & & 8B & 53.8 & 60.4 & 27.5 & 37.5 & 40.2 & 43.9 & 117.3\% & 27.3\% / 16.2\% / 56.6\% \\
\midrule
\multirow{9}{*}{\mistral}
    & No Adaptation & - & - & 37.6 & 33.5 & 28.0 & 15.0 & 23.8 & 27.6 & -- & 100\% / -- / --\\
    & RAG           & - & - & 30.2 & 28.1 & 35.0 & 16.0 & 25.8 & 27.0 & 97.9\% & -- / 100\% / -- \\
    & TTT           & - & - & 38.4 & 38.4 & 32.0 & 16.5 & 24.0 & 29.9 & 108.2\% & -- / -- / 100\%\\
    \cdashline{2-12}
    &   \multirow{4}{*}{\proj}  & \multirow{2}{*}{5.0} & 0.6B & 32.4 & 37.8 & 34.5 & 21.0 & 23.2 & 29.8 & 107.9\% & 4.6\% / 32.2\% / 63.2\%\\
    &   & & 8B & 31.4 & 34.8 & 32.0 & 22.5 & 28.0 & 29.7 & 107.8\% & 38.1\% / 31.5\% / 30.4\% \\
    \cdashline{3-12}
    &     & \multirow{2}{*}{8.0} & 0.6B & 32.8 & 37.8 & 37.5 & 19.5 & 23.3 & 30.2 & 109.4\% & 1.4\% / 33.1\% / 65.5\% \\
    &   & & 8B & 31.0 & 36.0 & 33.5 & 20.5 & 24.8 & 29.2 & 105.7\% & 22.4\% / 34.0\% / 43.6\% \\
    \cdashline{2-12}
    &  \multirow{4}{*}{\proj-Joint}   & \multirow{2}{*}{5.0} & 0.6B & 35.2 & 36.0 & 31.5 & 24.0 & 24.3 & 30.2 & 109.5\% & 4.6\% / 35.0\% / 60.4\% \\
    &   & & 8B & 31.4 & 36.6 & 34.0 & 21.0 & 27.0 & 30.0 & 108.7\% & 38.1\% / 28.9\% / 33.0\% \\
    \cdashline{3-12}
    &     & \multirow{2}{*}{8.0} & 0.6B & 33.8 & 36.6 & 39.0 & 20.5 & 22.7 & 30.5 & 110.6\% & 1.4\% / 32.9\% / 65.8\% \\
    &   & & 8B & 31.4 & 36.6 & 35.0 & 21.0 & 26.2 & 30.0 & 108.8\% & 22.4\% / 34.2\% / 43.4\% \\
\midrule
\multirow{9}{*}{\qwen}
    & No Adaptation & - & - & 41.2 & 26.2 & 26.0 & 32.5 & 26.5 & 30.5 & -- & 100\% / -- / --\\
    & RAG           & - & - & 38.8 & 41.5 & 32.5 & 42.0 & 26.3 & 36.2 & 118.8\% & -- / 100\% / -- \\
    & TTT           & - & - & 42.4 & 25.6 & 30.5 & 37.0 & 26.3 & 32.4 & 106.2\% & -- / -- / 100\% \\
    \cdashline{2-12}
    &  \multirow{4}{*}{\proj}   & \multirow{2}{*}{5.0} & 0.6B & 46.6 & 42.7 & 31.5 & 43.5 & 27.2 & 38.3 & 125.6\% & 12.1\% / 24.0\% / 63.9\% \\
    &   & & 8B & 48.6 & 45.1 & 27.0 & 41.5 & 27.0 & 37.8 & 124.1\% & 52.7\% / 20.6\% / 26.7\% \\
    \cdashline{3-12}
    &     & \multirow{2}{*}{8.0} & 0.6B & 46.0 & 39.0 & 31.0 & 50.0 & 25.0 & 38.2 & 125.3\% & 4.5\% / 25.4\% / 70.1\% \\
    &   & & 8B & 48.6 & 42.7 & 28.5 & 44.0 & 27.7 & 38.3 & 125.6\% & 35.8\% / 24.0\% / 40.2\% \\
    \cdashline{2-12}
    &   \multirow{4}{*}{\proj-Joint}  & \multirow{2}{*}{5.0} & 0.6B & 46.4 & 45.1 & 31.0 & 49.5 & 26.0 & 39.6 & 129.9\% & 12.1\% / 21.6\% / 66.4\% \\
    &   & & 8B & 46.8 & 46.3 & 27.5 & 45.5 & 26.0 & 38.4 & 126.1\% & 52.7\% / 16.7\% / 30.7\% \\
    \cdashline{3-12}
    &     & \multirow{2}{*}{8.0} & 0.6B & 46.4 & 43.9 & 32.5 & 51.5 & 26.2 & 40.1 & 131.5\% & 4.5\% / 24.7\% / 70.8\% \\
    &   & & 8B & 49.6 & 43.9 & 29.0 & 45.0 & 27.2 & 38.9 & 127.7\% & 35.8\% / 22.4\% / 41.8\% \\
\bottomrule
\end{tabular}
\caption{
\textbf{Impact of Reward Model in \proj.}
Performance comparison of adaptation strategies under varying reward models. ``0.6B'' refers to the \skyworkzerosixb, and ``8B'' refers to the \skyworkeightb. The ``Threshold ($\tau_r$)'' column denotes the minimum reward score required for direct response acceptance. ``Impr.'' reports relative improvement over the No Adaptation baseline. ``Strategy Distribution (\%)'' indicates the proportion of queries handled by No Adaptation, RAG, and TTT branches, respectively.
}
\label{tab:rm_analysis}
\end{table*}

\section{Analysis of Reward Model}
\label{sec:appendix_rm}

Table~\ref{tab:rm_analysis} presents a comprehensive evaluation of adaptation strategies across different LLMs, reward thresholds, and reward models. Here, ``0.6B'' denotes the \skyworkzerosixb reward model, and ``8B'' denotes the \skyworkeightb reward model.

\proj relies critically on the reward model to guide adaptive strategy selection. We investigate the impact of increasing reward model size by comparing the 0.6B and 8B variants. As shown in the table, using a larger reward model (8B) does not consistently yield significant improvements in accuracy across tasks. However, the 8B model tends to assign higher reward scores, resulting in a greater proportion of queries being handled by No Adaptation at the same threshold compared to 0.6B reward model. This shift indicates improved efficiency, as fewer queries require costly adaptation.
These findings highlight the necessity of tuning the reward threshold for each reward model to achieve optimal performance and efficiency in specific deployment scenarios. Additionally, it is important to consider that larger reward models incur higher inference costs, although this effect is negligible in the overall pipeline.

\end{document}